 \documentclass[final,5p,times,twocolumn,square,comma,numbers]{elsarticle}

\usepackage{amssymb}
\usepackage{lipsum}

\usepackage{amsmath,amsfonts}
\usepackage{algorithmic}
\usepackage{algorithm}
\usepackage{array}
\usepackage{textcomp}
\usepackage{stfloats}
\usepackage{url}
\usepackage{verbatim}
\usepackage{graphicx}
\usepackage{threeparttable}
\usepackage{diagbox}
\usepackage{pifont}
\usepackage{multirow}
\usepackage{bm}
\usepackage{tabularx}
\usepackage[misc]{ifsym}
\usepackage{caption}
\usepackage{subcaption}
\usepackage[colorlinks=true]{hyperref}
\usepackage{balance}
\usepackage{natbib}

\begin{document}

\begin{frontmatter}

\title{Bridging the Semantic-Numerical Gap: A Numerical Reasoning Method of \\ Cross-modal Knowledge Graph for Material Property Prediction}

\author{Guangxuan Song\textsuperscript{a},
Dongmei Fu\textsuperscript{a,\Letter},
Zhongwei Qiu\textsuperscript{c,d}, 
Zijiang Yang\textsuperscript{a}, \\
Jiaxin Dai\textsuperscript{a},
Lingwei Ma\textsuperscript{b},
and Dawei Zhang\textsuperscript{b,\Letter}
}

\affiliation{organization={Beijing Engineering Research Center of Industrial Spectrum Imaging, School of Automation and Electrical Engineering, University of Science and Technology Beijing},
            city={Beijing},
            country={China}
            }

\affiliation{organization={National Materials Corrosion and Protection Data Center, University of Science and Technology Beijing},
            city={Beijing},
            country={China}
            }

\affiliation{organization={Alibaba DAMO Academy},
            city={Hangzhou},
            country={China}
            }

\affiliation{organization={Zhejiang University},
            city={Hangzhou},
            country={China}
            }

\begin{abstract}
Using machine learning (ML) techniques to predict material properties is a crucial research topic. These properties depend on numerical data and semantic factors. Due to the limitations of small-sample datasets, existing methods typically adopt ML algorithms to regress numerical properties or transfer other pre-trained knowledge graphs (KGs) to the material.
However, these methods cannot simultaneously handle semantic and numerical information.
In this paper, we propose a numerical reasoning method for material KGs (NR-KG), which constructs a cross-modal KG using semantic nodes and numerical proxy nodes. It captures both types of information by projecting KG into a canonical KG and utilizes a graph neural network to predict material properties.
In this process, a novel projection prediction loss is proposed to extract semantic features from numerical information. NR-KG facilitates end-to-end processing of cross-modal data, mining relationships and cross-modal information in small-sample datasets, and fully utilizes valuable experimental data to enhance material prediction.
We further propose two new High-Entropy Alloys (HEA) property datasets with semantic descriptions.
NR-KG outperforms state-of-the-art (SOTA) methods, achieving relative improvements of 25.9\% and 16.1\% on two material datasets.
Besides, NR-KG surpasses SOTA methods on two public physical chemistry molecular datasets, showing improvements of 22.2\% and 54.3\%, highlighting its potential application and generalizability.
We hope the proposed datasets, algorithms, and pre-trained models$^*$ can facilitate the communities of KG and AI for materials.
\end{abstract}


\end{frontmatter}

\section{Introduction}
\label{introduction}
With the rise of big data in scientific and material research, the fourth paradigm of material data-driven research \cite{himanen2019data} is attracting widespread attention. In this context, machine learning (ML), with its outstanding modeling capabilities and reduced experimental costs, enables immense opportunities for
predicting \cite{ma2023data, dai2022cross} and exploring \cite{chan2022application, tawfik2020predicting} new materials based on their properties. 
However, ML in material science faces two challenges despite its achievements. 
First, high-quality material data is scarce due to the heavy reliance on experiments to obtain material properties. Material preparation and testing require substantial time and resources, leading to small data volume \cite{gao2022innovative}.
Second, effectively representing and utilizing multi-modal data is difficult \cite{wei2019machine}. Cross-modal ML requires a large amount of data, which is hard to obtain in material science. This dilemma often leads to the neglect of semantic information, such as processing techniques for material property prediction.

\begin{figure}[th!]  
    \centering
    \includegraphics[width=1.0\columnwidth]{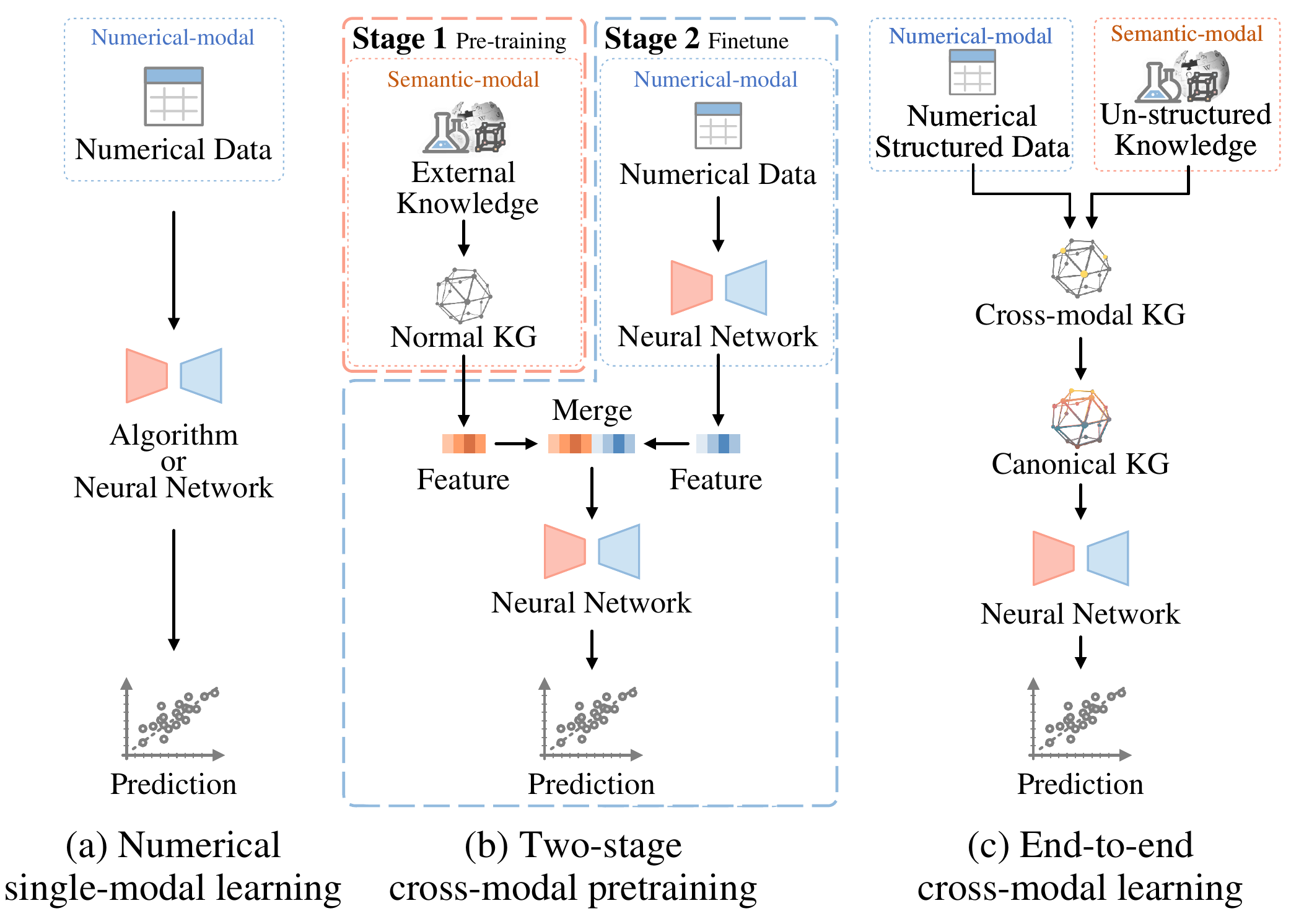}  
    \caption{Comparison of property prediction methods. The methods in (a) ignore material semantic information. The methods in (b) lack integration of numerical and semantic aspects. Our method in (c) utilizes numerical and semantic information while accounting for inter-sample relationships in small-sample data.}
    \label{fig1}
    \vspace{-0.3 cm}
\end{figure}

Two main approaches address the material property prediction challenges arising from the complexities of small-sample data and multi-modal data representation and utilization. 
Fig. \ref{fig1} (a) shows the numeric modality ML methods comprising classical ML methods \cite{li2022compressive, jahed2020examining} (e.g. random forests, RF) and neural network methods \cite{feng2019using}, based on material mechanisms and data features. 
These methods aim to uncover relationships between material numerical features and properties. They have advantages in small-sample problems but often overlook semantic information. 
Fig. \ref{fig1} (b) illustrates cross-modal pre-training methods that incorporate supplementary information, including external knowledge \cite{fang2023knowledge, fang2022molecular}. 
Pre-training with knowledge graphs (KGs) enhances ML by adding semantic knowledge, enabling better insights from small-sample data. 
KGs \cite{google2012} efficiently model semantic information and organize knowledge. 
However, constructing efficient domain-specific KGs demands extensive data and research efforts.
Furthermore, due to the limitations of two-stage methods, the absence of numerical data in the initial pre-training stage and the sole extraction of numerical features in the second numerical modeling stage limit the interaction between numerical and semantic information in feature extraction, resulting in a semantic-numerical gap.
The reason KGs are often used to handle semantic data alone is that normal KG construction primarily depends on text, and KG reasoning is primarily centered around modeling the KG's graph structure. 
Although certain research \cite{kristiadi2019incorporating, wu2018knowledge} has focused on KG representation learning involving numerical attributes, their modeling of numerical attributes aims to enhance node or relationship representations, effectively aligning numerical values with the semantic space.
While this enhances tasks such as link prediction, it still lacks the ability to effectively capture relationships between numerical values or the interactions between numerical and semantic information.
In scientific domains that require quantitative descriptions, especially in material property prediction problems, conducting numerical reasoning on cross-modal KGs becomes a focal point of our attention.
Simultaneously, both of the mentioned approaches in Fig. (a) and (b) suit small-sample material property prediction data. However, they exclusively focus on the relationship between material features and property, overlooking the relationship between samples in small-sample datasets.

To address the above problems, we propose the \textbf{N}umerical \textbf{R}easoning method of \textbf{K}nowledge \textbf{G}raph (\textbf{NR-KG}), as shown in Fig. \ref{fig1} (c). 
Inspired by KG pre-training and the incorporation of numerical attributes into KG representation learning, we aim to introduce numerical information into KG construction for forming a cross-modal KG. 
Initially, NR-KG extracts numerical and semantic information from material data and creates a cross-modal KG. 
Subsequently, we propose a novel cross-modal KG projection learning method, projecting the cross-modal KG into a canonical KG. 
To bridge the semantic-numerical gap in cross-modal projection, NR-KG uses a comparative learning loss for unified cross-modal learning and proposes an innovative projection prediction loss to supervise the learning of projection features in cross-modal information.
Ultimately, using graph neural networks (GNN), NR-KG propagates node information on the canonical KG based on real associations between materials, inferring material property from the material proxy node feature. 
The cross-modal KG structures diverse data with different processing or experimental conditions. 
Utilizing the projection learning of canonical KG, we fuse and model the semantic-numerical cross-modal information, establishing inter-sample relationships. 
NR-KG resolves the mentioned cross-modal representation and overlooked inter-sample relationship concerns. By comprehensively utilizing data, it enhances property prediction in small-sample scenarios without requiring extensive external knowledge pre-training.

Additionally, we propose two high-entropy alloys (HEA) datasets. HEAs are acclaimed in the scientific community for their exceptional mechanical properties, corrosion resistance, etc \cite{george2019high}. These datasets will facilitate research on ML techniques and new materials.

NR-KG is assessed using the HEA datasets. Moreover, NR-KG exhibits impressive performance in predicting molecular properties within the public dataset for scientific domain physical chemistry, underscoring its considerable potential in scientific data.

NR-KG is the first end-to-end KG numerical reasoning method, providing new insight into processing valuable experimental data for small-sample scientific data. Our contributions can be summarized as follows:

\begin{itemize}
    \item We propose two HEA datasets on hardness and corrosion resistance, integrating existing public HEA datasets and adding HEA data and semantic information from literature, offering diverse data.
    \item Our novel NR-KG method merges numerical and semantic cross-modal data, pioneering an end-to-end KG approach for semantic-numerical data mining.
    \item We propose an innovative projection prediction loss using high-dimensional generalized F-point to address the semantic loss issue in numerical information encoding, effectively capturing numerical-semantic information.
    \item Extensive experiments show that NR-KG achieves state-of-the-art results and demonstrates excellent application scalability on two datasets related to High Entropy Alloys (HEA) in materials science and two publicly available molecular property datasets in the scientific domain. Ablation studies emphasize the effectiveness of each model component.
\end{itemize}

\section{Related Work}

In addressing the material property prediction problem, we explored cross-modal knowledge graphs to enhance prediction accuracy in small-sample scenarios. Here, we provide a brief overview of related works on these issues.

\begin{figure*}[t]
\centering
\includegraphics[width=2.0\columnwidth]{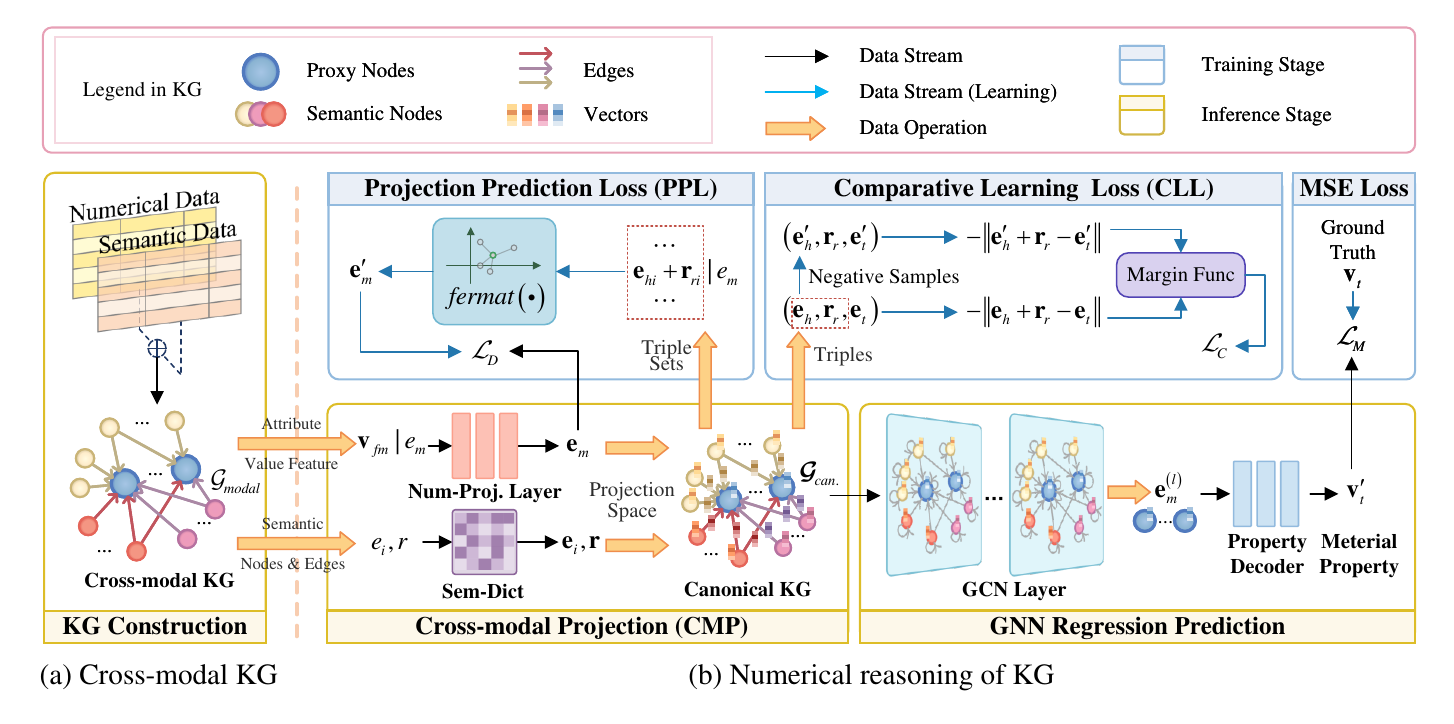} 
\caption{The NR-KG framework includes (a) KG construction and (b) numerical reasoning process. First, cross-modal KG $\mathcal{G}_{m}$ is constructed with numerical and semantic data. Then, proxy nodes represented by material numerical features $\mathbf{v}_{fm}|e_m$, semantic nodes $e_i$, and edges $r$ in $\mathcal{G}_{m}$ are projected into canonical KG $\bm{\mathcal{G}}_{c}$ using CMP. $\bm{\mathcal{G}}_{c}$ is supervised by CLL and PPL. GNN Regression Prediction facilitates information exchange and predicts material properties $\mathbf{v}'$. $\bigoplus$, Num-Proj. Layer, Sem-Dict, Margin Func, and $fermat(\cdot)$ refer to the join operation, Numerical Projection Layer, Semantic Dictionary, Margin function, and high-dimensional generalized F-point calculation.}
\label{fig2}
\end{figure*}

\subsection{Material Property Prediction}
The issue of utilizing machine learning (ML) for predicting material properties has been studied for decades. Due to the scarcity of material data and the high-dimensional complexity of material property spaces, researchers have focused on modeling problems with small-sample material data. 
Classical ML methods often outperform more complex deep learning networks on small-sample materials property prediction \cite{liu2022material}, due to their better generalization performance.
Expert prior knowledge enhances ML predictive prowess by formulating material descriptors with authentic physical causal relationships \cite{ghiringhelli2015big}, yielding valuable insights for small-sample ML.
\cite{zhang2022composition} introduces elemental information as external knowledge to enhance model predictive capabilities.
While more efficient deep learning feature extractors, like Graph Neural Networks (GNN) \cite{niu2024srr, yu2024identifying} or transformers \cite{wang2021compositionally}, have been applied to material property prediction tasks, capturing complex features, they come with increased data requirements. However, such methods encounter difficulties in integrating semantic information into modeling.

\subsection{KG Representation and Cross-modal Learning}
KG representation learning captures low-dimensional representations of entities and relationships with their semantics \cite{fionda2020learning, song2022knowledge} and has certain extrapolation abilities \cite{li2022does}.
Traditional KG representation learning methods, such as TransE \cite{bordes2013translating} and other translation-based models, excel in representing semantic information, but their ability to represent numerical properties is limited. 
These methods can flexibly encode semantic information, showing potential applications in scientific fields such as medicine \cite{zhu2020latte}. 
Recent research \cite{cvetkov2023relational, bayram2021node, kristiadi2019incorporating} tries to model numerical attributes in KGs to enhance semantic representation and improve link prediction accuracy. However, these methods mainly align numerical attributes (e.g. city latitude and longitude) embedding with the semantic embedding space, and do not address the problem of mining complex relationships between numerical and semantic values. 
End-to-end KG representation learning for joint reasoning of numerical and semantic information in scientific domains is still relatively limited.

\subsection{Small-sample Learning}

In materials science research, scenarios with limited sample data are extremely common. The challenge lies in capturing the crucial features of the data within these limited samples to avoid overfitting during training.
Transfer learning, involving the transfer of knowledge from one domain or task (source) to another (target) \cite{pan2009survey}, reduces the data requirements for the target domain and has mature applications. These methods can provide effective prior knowledge but necessitate a reasonably close distribution match between pre-training and target datasets. Additionally, during the pre-training phase, appropriate tasks need to be designed to capture features that generalize well for downstream tasks.
Recent work has employed transfer learning strategies to introduce external knowledge and enhance predictive performance. In the \cite{fang2023knowledge, fang2022molecular}, knowledge was pre-trained on an ElementKG, and then this pre-trained external knowledge was incorporated into the process of predicting molecular properties. 
Nevertheless, these methods demand substantial data and computational resources, and the two-stage process creates a numerical-semantic gap, not fully leveraging the limited data. Furthermore, the potential value of inter-sample correlation information remains underexplored.

\section{Method}

To address the challenges of modeling semantic information and bridging the numerical-semantic gap in scientific small-sample data scenarios, as well as establishing relationships among samples and making optimal use of limited data, we have defined the tasks of material property prediction, cross-modal KG numerical reasoning and numerical reasoning projection learning in Sec. \ref{m.1}. The NR-KG framework is proposed in Sec. \ref{m.2}, with technical details presented in Sec. \ref{m.3}, \ref{m.4}, and \ref{m.5}. Sec. \ref{m.6} provides the complete algorithm for NR-KG.

\subsection{Problem Formulation\label{m.1}}
\subsubsection{Material Property Prediction}

Material property prediction forecasts the physical and chemical characteristics of materials through a regression model that uses known features and their relationships. This process is mathematically represented as
\begin{equation}
\label{eq:material_property_prediction}
v = F_{\text{property}}(\mathcal{X}_p, \mathcal{R}_p),
\end{equation}
where $v \in \mathbb{R}$ is the predicted material property value, $\mathcal{X}_p$ includes physical and chemical features, among other related information, $\mathcal{R}_p$ encompasses the relationships between these features, and the function $F_{\text{property}}$ is the material property prediction model.

Unlike common sense KG numerical predictions, material property predictions are constrained by fundamental physical laws of the scientific world. The data-driven prediction process aims to fit complex physical models using data. In contrast, the numerical information in common sense KG numerical attribute (like latitude, longitude, year, etc.) modeling represents more semantic information, serving as context distinct from discrete semantic information (nodes or edges). Therefore, when considering the integration of semantic information, the numerical and semantic information in material property prediction differs in form and content across modalities.

\subsubsection{Numerical Reasoning on Cross-modal KGs}
KG is a knowledge model that structurally describes knowledge using nodes and edges. Let there be a node set $\mathcal{E}$, an edge set $\mathcal{R}$, an attribute set $\mathcal{A}$, and a real number set $\mathcal{V}$. Then, the cross-modal KG ($\mathcal{G}_{m}$) is represented as a set of triples 
\begin{equation}
\label{eq:g}
    \begin{split}
        \mathcal{G}_{m} = \{( e_1,r,e_2 ) \} \cup \{( e,a,v ) \},
    \end{split}
\end{equation}
where $(e_1,r,e_2) \in \mathcal{E} \times \mathcal{R} \times \mathcal{E}$ , $(e,a,v) \in \mathcal{E} \times \mathcal{A} \times \mathcal{V}$. In the attribute set $\mathcal{A}$, we define $\mathcal{A}_f\cup \mathcal{A}_{t}=\mathcal{A}$, where the subset $\mathcal{A}_f$ represents the feature attribute set that describes the features of relevant entities, while $\mathcal{A}_{t}$ represents the target attribute set that describes the properties of entities.

Numerical reasoning constructs two function mappings, $\psi: \mathcal{V} \to \mathcal{E}$ and $\Omega: \mathcal{E} \times \mathcal{R} \times \mathcal{E} \to \mathcal{V}$, which predict the target attributes of entities based on their feature attributes and triple information.

\subsubsection{Numerical Reasoning Projection Learning}
Let us define the projection dimension as $H$. We define the function ${f}:\mathcal{V}\to {{\mathbb{R}}^{H}}$ to map the node's feature attribute $\mathbf{v}$ to its canonical vector ${f} ( \mathbf{v} )$, and the function ${g}:{{\mathbb{R}}^{H}}\times {{\mathbb{R}}^{H}}\times {{\mathbb{R}}^{H}}\to \mathcal{V}$ to map the triple information $\left( \mathbf{e}_1,\mathbf{r},\mathbf{e}_2 \right)$ to a vector relevant to the target attribute value. Thus, numerical reasoning projection learning on $\mathcal{G}_{m}$ is defined as $\psi( v ) \xrightarrow{\mathrm{def}} {f}( \mathbf{v} )$ and $\Omega ( e_1,r,e_2 )\xrightarrow{\mathrm{def}}{g}( \mathbf{e}_1,\mathbf{r},\mathbf{e}_2 )$.

\subsection{Framework\label{m.2}}

The structure of \textbf{NR-KG} is shown in Fig. \ref{fig2}.
A cross-modal KG $\mathcal{G}_{m}$ is constructed through the Cross-modal KG Construction process. 
Then, $\mathcal{G}_{m}$ is projected into the canonical KG, denoted as $\bm{\mathcal{G}}_{c}$, by the Cross-modal Project (CMP). The supervision involves two primary components: Projection Prediction Loss (PPL) and Comparative Learning Loss (CLL). 
Lastly, GNN Regression Prediction is utilized for the forecast of properties represented by material proxy nodes. 
Notably, our framework is designed as an end-to-end process, fostering the integration of numerical and semantic data within the ($\mathcal{G}_{m}$). 
NR-KG approach ensures that the model captures intricate relationships and patterns in the data, leading to more accurate predictions and a deeper understanding of material properties.

\subsection{Cross-modal KG Construction\label{m.3}}

We classify material data into two types: numerical data (e.g., material composition) and semantic data. The latter encompasses text-based descriptions of material manufacturing processes, structures, or classification information from various perspectives. Given the flexible nature of semantic data, which is influenced by recorder subjectivity and the presence of multiple process combinations or missing information, a cross-modal KG is proposed to capture the versatile relationships between numerical and semantic data in material datasets.

The process of cross-modal KG ($\mathcal{G}_{m}$) construction is outlined as follows: 
To incorporate numerical data, we set a proxy node ${e}_{m}$ for each material. 
Numerical attributes characterizing material properties are subsequently added as node feature attributes, denoted as a triple set $\{({{e}_{m}},{{r}_{f}},{{v}_{fm}})\}$, where ${{r}_{f}}$ and ${{v}_{fm}}$ denote the feature attribute and the numerical feature value of the material, respectively.
These numerical feature value represent the independent variables that exert influence on the properties of the material. 
Moreover, numerical data describing material properties is added as target attribute values for ${e}_{m}$, denoted as a triple set $\{({{e}_{m}},{{r}_{t}},{{v}_{tm}})\}$, where ${{r}_{t}}$ and ${{v}_{tm}}$ correspond to the target attribute and numerical target attribute value of ${e}_{m}$. These values correspond to the dependent variables interlinked with the properties of the material.

Concerning semantic information, we extract named entities and treat them as separate nodes. 
These nodes are then linked to relevant material proxy nodes, creating semantic associations denoted as $( {{e}_{m}},{{r}_{k}},{{e}_{i}} )$ or $( {{e}_{i}},{{r}_{l}},{{e}_{j}} )$, where $e_i$ and $e_j$ signify different nodes, and ${r}_{k}$ and ${{r}_{l}}$ represent different edges within $\mathcal{G}_{m}$.
Additionally, We perform an operation 
\begin{equation}
    ( {{e}_{i}},{{r}_{k}},{{e}_{m}} )\leftarrow ( {{e}_{m}},{{r}_{k}},{{e}_{i}} ),
\end{equation}
to reverse relationships originating from material nodes. This aligns with the GNN Regression Prediction in NR-KG, transforming relationship descriptions from the ``active voice'' to the ``passive voice''.

Cross-modal KG Construction introduces material proxy nodes as repositories of numerical data. This effectively organizes data in numerical-modal and semantic-modal forms, which lack an inherent structure, into a cohesive KG.

\subsection{Cross-modal KG Projection to Canonical KG\label{m.4}}

Cross-modal Projection Module maps each node and edge in the cross-modal KG ($\mathcal{G}_{m}$) to a projection space $\mathbb{R}^{H}$, referred to as the canonical KG ($\bm{\mathcal{G}}_{c}$), which plays a pivotal role in bridging the semantic-numerical gap. The projection dimension is denoted by $H$.
For the semantic nodes $e_i$ and edges $r$, they are independently mapped to vectors $\mathbf{e}_i \in \mathbb{R}^{H}$ and $\mathbf{r} \in \mathbb{R}^{H}$ by a semantic dictionary:
\begin{equation}
\label{eq:semd}
\begin{array}{c}
\begin{aligned}
    \mathbf{e}_i & = Norm \left( f_{SemD}\left({e}_i; E_s\right) \right) \\
    \mathbf{r} & = f_{SemD} \left(r; E_s \right)    
\end{aligned}
\end{array},
\end{equation}
where $E_s$ denotes the learnable parameters, function $f_{SemD}$ represents the mapping of the semantic dictionary, and $Norm$ represents the normalization operation to ensure $\mathbf{e}_i$ is within a reasonable range, stabilizing model training.
For the proxy nodes $e_m$, we propose the numerical projection layer to predict their vectors $\mathbf{e}_m$ in $\bm{\mathcal{G}}_{c}$ based on the vector of node feature attributes $\mathbf{v}_{fm}$. This ensures that the projection vector of $e_m$ is independent of the semantic dictionary. 
The implementation of the numerical projection layer employs the multi-layer perceptron (MLP) network in NR-KG, as MLP can preserve numerical features:
\begin{equation}
\label{eq:npl}
    \mathbf{e}_m = Norm\left(f_{NPL}\left(\mathbf{v}_{fm}; \theta_{NPL}\right)\right),
\end{equation}
where $f_{NPL}$ is the implementation of the numerical projection layer, and $\theta_{NPL}$ is network parameters.

To ensure that $\bm{\mathcal{G}}_{c}$ after cross-modal projection can simultaneously fuse numerical and semantic information, we were inspired by the translation-based method TransE \cite{bordes2013translating} to design the comparison learning loss (CLL). This loss assumes that $\mathbf{e}_h$ can be translated from $\mathbf{e}_t$ based on $\mathbf{r}_r$ in the triple $({e}_h, {r}_r, {e}_t)$.
For the triples $(e_m, r_k, e_i)$ or $(e_i, r_l, e_j)$ in $\mathcal{G}_{m}$, we uniformly refer to them as $(e_h, r_r, e_t)$. The CLL defines a distance function $d(e_h, r_r, e_t) = -\| \mathbf{e}_h + \mathbf{r}_r - \mathbf{e}_t \|$ to evaluate the plausibility of the triple $(e_h, r_r, e_t)$, where $\|\cdot\|$ represents the L2-norm.
The goal of comparative learning within $\bm{\mathcal{G}}_{c}$ is to minimize a margin loss, given by:
\begin{equation}
    \mathcal{L}_C\!=\!
    \begin{cases}
    d(e_h, r_r, e_t)\!\!&\!,(e_h, r_r, e_t) \! \in \! \mathcal{G}_{m}\\
    \max (0, \gamma - d(e_h', r_r, e_t'))\!\!&\!,(e_h', r_r, e_t') \! \notin \! \mathcal{G}_{m}
    \end{cases},
    \label{eq:lc}
\end{equation}
where $\gamma > 0$ serves as a hyperparameter that differentiates between positive and negative samples, $(e_h', r_r, e_t')$ denotes negative samples extracted from $\mathcal{G}_{m}$.
A feasible method for constructing these negative samples involves randomly substituting the $e_h$ and $e_t$ in the positive triples $(e_h, r_r, e_t)$.

Predicting the projection vector $\mathbf{e}_m$ by learning the numerical data of the proxy node ${e}_m$ remains a challenge, primarily due to the common situation where the projection dimension $H$ is much greater than the dimension of $\mathbf{v}_{fm}$. Additionally, there's a lack of semantic information during the process of predicting $\mathbf{e}_m$ from $\mathbf{v}_{fm}$ using the numerical projection layer.
Hence, we propose a reverse method that involves directly computing the projection vector of $e_m$ using the concept of CLL. This method is termed projection prediction loss (PPL) and serves to enhance the supervision of the numerical projection layer in learning $\bm{\mathcal{G}}_{c}$ predictions.

PPL is built on a fundamental assumption from CLL: 
$\mathbf{e}_h + \mathbf{r}_r \approx \mathbf{e}_t$. As we place $e_m$ in the tail entity slot while constructing $\mathcal{G}_{m}$,
the computed value for the specific $\mathbf{e}_m$ in the $\bm{\mathcal{G}}_{c}$ is defined as an optimization problem:

\begin{equation}
{\mathbf{e}_m' = 
\underset{\mathbf{e} \in \mathbb{R}^{H}}{\arg\min} \sum_{\mathbf{p} \in \mathcal{P}} \left\| \mathbf{e} - \mathbf{p} \right\|},
\end{equation}

where $\mathcal{P}$ means all sets of $\mathbf{e}_h + \mathbf{r}_r$ in $\{(\mathbf{e}_h, \mathbf{r}_r, \mathbf{e}_t)\}$ for a specific $\mathbf{e}_t$.
Thus, our objective is to find a point among all points $\mathbf{e}_m'$ that minimizes the sum of distances of all points within $\mathcal{P}$. This point $\mathbf{e}_m'$ corresponds to a high-dimensional generalized F-point \cite{li2022cohesive}, which can be solved through optimization methods. 
Due to the geometric complexity of high-dimensional space, the high-dimensional generalized F-point typically lacks an explicit solution, and numerical optimization methods are used to obtain its solution within an acceptable range. 
In NR-KG, for computational efficiency, we employed the average point of $\mathbf{p}$ to estimate $\mathbf{e}_m'$.

To learn the projection vector $\mathbf{e}_m$ of proxy nodes, PPL minimizes a distance loss given by:
\begin{equation}
    {\mathcal{L}_D = \sum\limits_{e_m \in \mathcal{E}} \left\| \mathbf{e}_m - \mathbf{e}_m' \right\|}.
\end{equation}
The losses $\mathcal{L}_C$ and $\mathcal{L}_D$ govern the process of cross-modal KG projection to the canonical KG, establishing connections between numerical and semantic information, ensuring alignment within the canonical KG. In the learning process of the canonical KG, the projection vector $\mathbf{e}_m$ for proxy nodes $e_m$ is defined as an encoding process that depends solely on numerical features. This process, compared to traditional KG embedding methods (e.g., \cite{bordes2013translating}), allows flexible incorporation of new proxy nodes during the inference stage, which holds practical significance for applications in the materials domain.

\subsection{GNN Regression Prediction\label{m.5}}

\noindent
The target features of proxy nodes depend not only on the proxy nodes themselves but also on their neighboring nodes. To model this relationship, we utilize the graph convolutional network (GCN) to propagate node feature vectors in accordance with real-world information relationship logic.
During this process, we disregard the relations vectors in canonical KG and treat them as connections. The GCN operation is as follows:
\begin{equation}
\label{eq_gnn}
    {\mathbf{e}}^{\left( l+1 \right)}=\sigma \left( {{{\tilde{D}}}^{-\frac{1}{2}}}\tilde{A}{{{\tilde{D}}}^{-\frac{1}{2}}}{{\mathbf{e}}^{\left( l \right)}}\mathbf{W} \right),
\end{equation}
where, $\tilde{A}=A+I$, which $A$ is the critical matrix, $\tilde{A}$ represents the addition of self-loops, and $\tilde{D}$ stands for the degree matrix of $\tilde{A}$, and $\sigma(\cdot)$ denotes the activation function.

Finally, we use an MLP network as a property decoder to predict the target attribute values $\mathbf{v}_t$ based on the proxy node features:
\begin{equation}
    \mathbf{v}'_t = f_{PD}\left(\mathbf{e}_m; \theta_{PD}\right),
\end{equation}
where function $f_{PD}$ is the property decoder, $\theta_{PD}$ is network parameters, and $\mathbf{v}'_t$ denotes the prediction of the target attribute values $\mathbf{v}_t$.

This step involves learning a mean squared error (MSE) loss, defined as 
\begin{equation}
\mathcal{L}_{M}=\sum\limits_{v_t\in V}{\left\| \mathbf{{v}'_t}-\mathbf{v_t} \right\|}, 
\end{equation}
where $\|\cdot\|$ denotes the L2-norm.

\begin{algorithm}[ht]
\caption{NR-KG Algorithm}
\label{alg:algorithm}

\textbf{Input}
$D$: dataset of material information.

\textbf{Parameter}
$\gamma$, $\gamma_a$, $\gamma_b$: the hyperparameter in $\mathcal{L}$.
$H$: canonical KG ($\bm{\mathcal{G}}_{c}$) dimension.
$T$: maximum iteration.

\textbf{Output}
$\mathbf{v}'_{t}$: predicted target attribute value for proxy node.

\textbf{Training stage}:
\begin{algorithmic}[1]
\STATE Construct a cross-modal KG $\mathcal{G}_{m}$ from $D$.
\STATE Initialize the semantic dictionary $E_s$
\FOR {$epoch = 1$ to $T$}
    \FOR {$(e_h, r_r, e_t) \in \mathcal{G}_{m}$}
        \FOR {$t \in (e_h, r_r, e_t)$}
            \STATE \textbf{if} $t \in \{e_m\}$ \textbf{then} $\textbf{t} \leftarrow Norm(f_{NPL}(v_{fm}))$
            \STATE \textbf{else} $\textbf{t} \leftarrow Norm(f_{SemD}(t, E_s)) \text{ or } f_{SemD}(t, E_s)$
        \ENDFOR  
    \ENDFOR
    \STATE $\mathcal{L}_C \leftarrow \sum \max(0, d(e_h', r_r, e_t') - d(e_h, r_r, e_t) + \gamma )$
    \STATE ${\mathcal{L}_D \leftarrow \sum \left\| \mathbf{e}_m - {\arg\min} \sum \left\| \mathbf{e} - \mathbf{p} \right\| \right\|}$
    \STATE $\bm{\mathcal{G}}_{c}$ $ \leftarrow f_{GCN}(G)$ \COMMENT{Ref Eq.(\ref{eq_gnn})}
    \STATE $\mathbf{v}'_{tm} \leftarrow f_{PD}(\mathbf{e}_m)$ \textbf{for} each $e_m$ in $\bm{\mathcal{G}}_{c}$
    \STATE $\mathcal{L}_{M} \leftarrow \sum{ \left\| \mathbf{{v}'_t}-\mathbf{v_t} \right\|}$
    \STATE Update the model w.r.t. $\mathcal{L} = \mathcal{L}_{M} + \gamma_a \mathcal{L}_{C} + \gamma_b \mathcal{L}_{D}$
\ENDFOR
\RETURN $\mathbf{v}'_{t}$
\end{algorithmic}

\textbf{Inference stage}:
\begin{algorithmic}[1]
\STATE $\mathcal{G}_{m} \xleftarrow{\mathrm{add}} \{(e_m, r_f, v_{fm})\}$
\STATE Same as lines 4 to 9, 12, 13 and 17 in the training stage
\RETURN $\mathbf{v}'_{t}$
\end{algorithmic}
\end{algorithm}

\subsection{Learning\label{m.6}}

\subsubsection{Algrithom}
The training and inference stages are shown in Algorithm \ref{alg:algorithm}.
In the training stage, during each iteration, the cross-modal KG is initially projected onto the canonical KG, followed by GCN calculations to determine the material property values of the material proxy nodes. Subsequently, the loss is calculated to update the network parameters. In the testing stage, after adding the predicted material entities and relevant relationships to the cross-modal KG, the material property can be obtained using the trained network.

\subsubsection{Loss Function}
The overall loss of NR-KG is denoted as 
\begin{equation}
    {\mathcal{L}={\mathcal{L}_{M}}+\gamma_a {\mathcal{L}_{C}}+\gamma_b {\mathcal{L}_{D}}},
\end{equation}
where ${\mathcal{L}_{M}}$ denotes the mean squared error loss, ${\mathcal{L}_{C}}$ denotes the comparative learning loss, and ${\mathcal{L}_{D}}$ denotes the projection prediction loss, $\gamma_a$ and $\gamma_b$ are coefficients of ${\mathcal{L}_{C}}$ and ${\mathcal{L}_{D}}$.

\section{Experiments}

To validate the performance, effectiveness, and generalizability of NR-KG, comprehensive experiments were conducted. Sec. \ref{E.1} details the experimental settings, focusing on the two new HEA datasets in Sec. \ref{E.1.1}. Sec. \ref{E.2} describes the construction process and statistical information of the cross-modal KG. Comparative experiments between NR-KG and state-of-the-art (SOTA) methods on our HEA datasets, highlighting NR-KG’s outstanding performance, are presented in Sec. \ref{E.3}. \ref{E.4} and \ref{E.5} respectively provide insights into ablation and interpretability experiments, elucidating NR-KG’s working principles. To demonstrate NR-KG’s application scalability, Sec. \ref{E.6} presents comparative experiments on two publicly available molecular datasets.

\subsection{Experiments Settings\label{E.1}}

\subsubsection{HEA Dataset Construction and Dataset Selection\label{E.1.1}}
The experimental dataset comprises two categories: two high-entropy alloy (HEA) datasets constructed by us to demonstrate the effectiveness of NR-KG in material property prediction tasks, and two publicly available physical chemistry molecule property datasets used to validate the application scalability of NR-KG.

HEA datasets comprise HEA public datasets and newly collected data. We focused on HEAs with Al-Fe-Ni-Co-Cr-Cu-Mn elements due to extensive research in materials science. We combined data from \cite{yang2022machine, nyby2021electrochemical, xiong2021machine}, adding semantic information (processing techniques and crystal structures) from the original literature. Additionally, we enriched the dataset with recent high-level publications in materials research from Scopus. During extraction, GPT-3.5 \cite{gpt35} aided in locating and extracting key information, verified and added by materials experts. Original expressions were retained without vocabulary alignment to assess method effectiveness in the presence of semantic noise(e.g. errors, omissions, ambiguities).

We propose the HEA hardness dataset (HEA-HD) and the HEA corrosion resistance dataset (HEA-CRD). All materials include elemental atomic ratios, while processing techniques and crystal structure information have some missing values. HEA-HD comprises 397 items labeled with HV (higher values indicating greater hardness). HEA-CRD contains 151 items labeled with logarithmically processed corrosion current $\ln(I_{corr})$ (lower values indicating better corrosion resistance). Details of data collection, cleaning, and a detailed dataset analysis are provided in \ref{apx_dataCollection}.

Due to limitations in previous algorithms in handling semantic information, most publicly available material datasets lack semantic data aligned with numerical information.
To validate the applicability of NR-KG, we conduct experiments on a public benchmark dataset  FreeSolv \cite{mobley2014freesolv} and ESOL \cite{delaney2004esol}. 
Prediction of molecular properties is a central research topic in physics, chemistry, and materials science that focuses on predicting various properties and characteristics of molecules based on their structural representations. 
Therefore, the task of predicting molecular properties is used to evaluate the applicability of NR-KG in the field of science.

\begin{figure*}[ht!]  
    \centering
    \includegraphics[width=1.9\columnwidth]{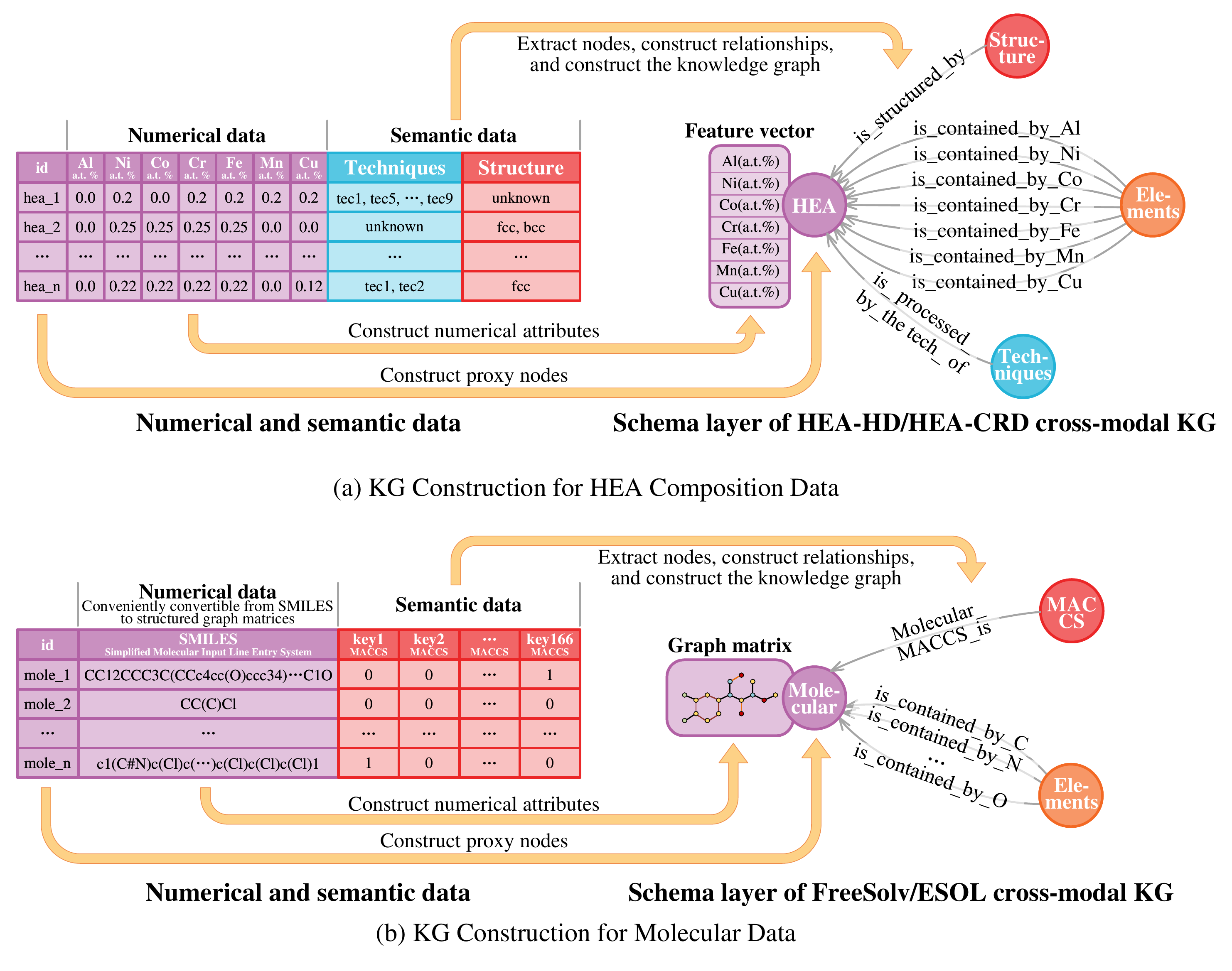}
    \caption{Cross-modal KG construction and schema layer for HEA and molecular data. (a) and (b) show the process of HEA and molecular data, respectively. In the tables of numerical and semantic data and in the schema layer of the cross-modal KG, the same colors indicate identical categories of information.}
    \label{fig3}
\end{figure*}

\subsubsection{Data Split and Evaluation Metrics}
For HEA-HD and HEA-CRD datasets, we adopted the 6-fold cross-validation method: The dataset was split into 6 equal parts, cyclically using 4 parts for training and the remaining 2 parts for validation and testing. During splitting, we ensured the training set included all semantic words to construct a complete semantic dictionary.

For FreeSolv and ESOL datasets, we apply the scaffold splitting \cite{bemis1996properties} method. Scaffold splitting splits the molecules with distinct two-dimensional structural frameworks into different subsets, providing a more challenging but practical setting since the test molecule can be structurally different from the training set. Following \cite{fang2022geometry, zhou2023uni, fang2023knowledge}, the data were split into training, validation, and testing sets at an 8:1:1 ratio using scaffold splitting. This random partitioning was repeated three times to ensure the statistical validity of the experimental results.

Two types of evaluation metrics were employed:
\begin{itemize}
    \item For HEA property prediction experiments (numerical regression task): mean squared error (MSE, smaller is better), mean absolute error (MAE, smaller is better), and coefficient of determination (R$^2$, larger is better). 
    Additionally, for comparability with public results, root mean square error (RMSE, smaller is better) is used to assess molecular property prediction performance.
    \item For model interpretability experiments (link prediction task): mean reciprocal rank (MRR, larger is better), mean rank (MR, lower is better), and Hits@1 (top one prediction hit rate, larger is better).
\end{itemize}

\subsection{Cross-modal KG Construction and Analysis\label{E.2}}
\noindent
We developed scripts to transform the datasets into triples (a structured KG format) as shown in Fig. \ref{fig3}. For HEA data, HEA-HD and HEA-CRD were processed into triples where different processing techniques, crystal structures, and elements are considered as different nodes. Each HEA is assigned a HEA proxy node, linked to processing techniques, crystal structures, and element nodes through relations in ``passive voice''. Element composition content become feature attribute values for proxy nodes, and property values are treated as target attribute values.

For molecular data, given the absence of direct semantic information in public datasets, we treated the inclusion relationships of molecular substructures as semantic data to test our NR-KG method. Additionally, we considered the molecular structure matrices—including transition matrices and node attribute matrices—as numerical data. This step is crucial for creating molecular proxy nodes. Defined by the Molecular ACCess System (MACCS) keys \cite{durant2002reoptimization} as semantic nodes, the 166 substructures facilitated connections between these molecular proxy nodes and semantic nodes based on their molecular MACCS fingerprints \cite{duvenaud2015convolutional}. This method quickly created semantic links between molecules and utilized the RDKit open-source toolkit \cite{landrum2013rdkit} for comprehensive molecular structure analysis.

\begin{table*}[t]
\caption{Scale of the cross-modal KGs. ``Unconnected HEA Nodes'' and ``Unconnected molecular nodes'' represent the count of HEA nodes and molecular nodes that are not linked to any processing, crystal structure, or MACCS nodes, respectively.}

\label{table1}
  \centering
  \begin{subtable}[h]{0.45\textwidth}
    \centering
    \caption{Cross-modal KGs of HEA datasets.}

        \resizebox{\linewidth}{!}{
        \begin{threeparttable}
        \renewcommand\tabcolsep{15pt}
        \begin{tabular}{c | cc}
        \hline
        
        \hline
        Items & HEA-HD & HEA-CRD \\ \hline
        HEA nodes & 397 & 151 \\ 
        Element nodes & 7 & 7 \\ 
        Processing techniques nodes & 33 & 51 \\ 
        Crystal structure nodes & 7 & 9 \\ 
        Relationship category & 9 & 9 \\ 
        Average degree of nodes    & 7.191 & 6.046 \\ 
        Unconnected HEA nodes & 194 & 15 \\ 
        \hline
        
        \hline
        \end{tabular}
        \end{threeparttable}
        }

  \end{subtable}
  \hspace{0.2 cm}
  \begin{subtable}[h]{0.45\textwidth}
    \centering
    \caption{Cross-modal KGs of molecular datasets.}
        \resizebox{\linewidth}{!}{
        \begin{threeparttable}
        \renewcommand\tabcolsep{15pt}
        \begin{tabular}{c | cc}
        \hline
        
        \hline
        Items & FreeSolv & ESOL \\
        \hline
        Molecular nodes & 642 & 1128 \\
        Element nodes & 9 & 9 \\
        MACCS nodes & 147 & 147 \\
        Relationship category & 10 & 10 \\
        Average degree of nodes & 28.048 & 45.925 \\
        Unconnected molecular nodes & 0 & 0 \\
        \hline
        
        \hline
        \end{tabular}
        \end{threeparttable}
        }
  \end{subtable}
\end{table*}

\begin{figure*}[!t]  
    \centering
    \includegraphics[width=2\columnwidth]{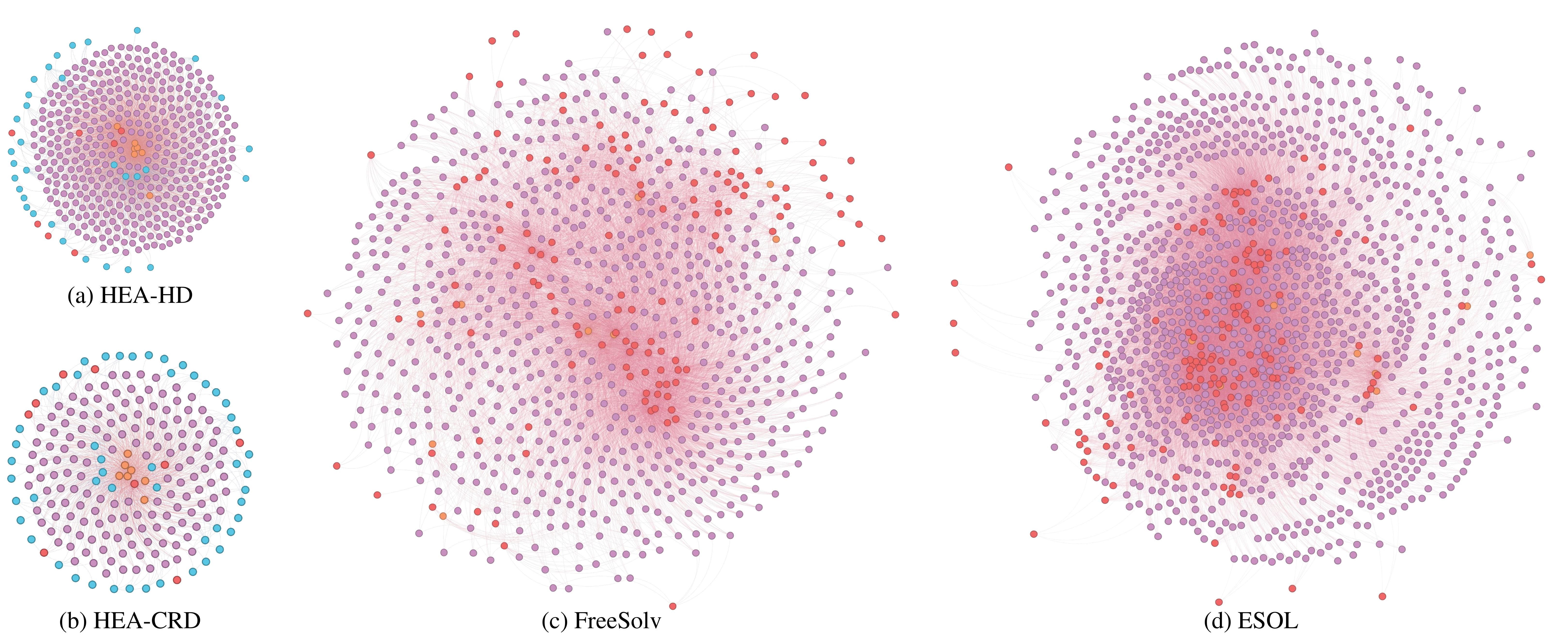}
    \caption{Visualization of cross-modal KGs. (a), (b), (c) and (d) depict the visualization of cross-modal KGs constructed from the HEA-HD, HEA-CRD, FreeSolv and ESOL datasets, respectively.}
    \label{fig_KGs}
\end{figure*}

Table \ref{table1} and Fig. \ref{fig_KGs} respectively show the statistical scale and visualization of cross-modal KGs. The HEA dataset is derived from real experiments in materials science publications. Consequently, some HEA nodes link to multiple processing techniques and crystal structure nodes, while others remain isolated. These cross-modal KGs effectively capture these complex relationships.

In the NR-KG training stage, HEA nodes in the validation and testing sets are temporarily masked to prevent their influence, forming the training cross-modal KG. These nodes are reintroduced during the validation and testing phases. However, some KG-based methods in comparison experiments cannot introduce new nodes during testing. 
For these methods, we only mask the target property of the HEA in the validation and testing sets, without deleting any nodes. This approach allows these methods to utilize semantic data from the validation and testing sets during training, thus reducing the difficulty of performance prediction tasks. 
Consequently, NR-KG tasks more closely reflect real-world scientific data prediction scenarios and pose greater challenges.

\subsection{Comparison Experiments\label{E.3}}

\begin{table*}[!t] 
\caption{Comparison experiment results. Results are shown as ``mean ± standard deviation''. Bold cells highlight the best results per metric. The ``New nodes'' column signifies the method's ability to predict new HEAs without retraining. CMM, IEKM, KG-NAMM, and KG-AM denote classical ML-based methods, incorporating element knowledge method, KG numerical attribute modeling methods, and KG augmentation method, respectively. Our method, NR-KG, achieved a 25.9\% and 16.1\% improvement over the second-best algorithm of MSE.}
\label{mianTable}
\centering
\begin{threeparttable}
\resizebox{\linewidth}{!}{
\begin{tabular}{@{} c|c|c | r@{±}l r@{±}l r@{±}l | r@{±}l r@{±}l r@{±}l @{}}
\hline

\hline
\multicolumn{3}{c|}{Datasets} & \multicolumn{6}{c|}{HEA-HD}                                                             & \multicolumn{6}{c}{HEA-CRD}                                                \\
\hline
\multicolumn{3}{c|}{Task}    & \multicolumn{6}{c|}{Regression (Testing set)}                                                            & \multicolumn{6}{c}{Regression (Testing set)}                                                           \\
\hline
Category & Methods & New nodes & \multicolumn{2}{c}{MSE(×10$^3$)$\downarrow$} & \multicolumn{2}{c}{MAE$\downarrow$} & \multicolumn{2}{c|}{R$^2$$\uparrow$} & \multicolumn{2}{c}{MSE$\downarrow$} & \multicolumn{2}{c}{MAE$\downarrow$} & \multicolumn{2}{c}{R$^2$$\uparrow$}\\
\hline
\multirow{6}{*}{CMM} & YSRF           & \checkmark          & 5.438          & 1.706          & 46.853          & 6.837          & 0.820          & 0.031          & 3.895          & 1.592          & 1.485          & 0.350          & 0.202          & 0.271          \\
 & YSRF(semantic) & \checkmark          & 4.825          & 1.604          & 43.932          & 7.564          & 0.841          & 0.036          & 3.015          & 1.260          & 1.313          & 0.246          & 0.362          & 0.278          \\
 & HDSVR          & \checkmark          & 6.441          & 1.312          & 63.555          & 3.958          & 0.783          & 0.039          & 5.039          & 2.149          & 1.747          & 0.281          & -0.064         & 0.362          \\
 & HDSVR(semantic) & \checkmark          & 5.773          & 1.238          & 61.066          & 6.175          & 0.804          & 0.045          & 2.816          & 1.110          & 1.284          & 0.152          & 0.408          & 0.184          \\
 & PCHMLP         & \checkmark          & 6.017          & 1.976          & 58.110          & 9.069          & 0.801          & 0.046          & 6.633          & 1.437          & 2.125          & 0.220          & -0.426         & 0.304          \\
 & PCHMLP(semantic) & \checkmark          & 5.557          & 1.483          & 55.417          & 7.371          & 0.810          & 0.057          & 2.633          & 0.986          & 1.161          & 0.146          & 0.449          & 0.168          \\
\hline
\multirow{1}{*}{IEKM} & DS-HEA      & \checkmark          & 6.632          & 2.018          & 62.633          & 13.743         & 0.770          & 0.093          & 5.502          & 0.853          & 1.809          & 0.200          & -0.096         & 0.300          \\
\hline
\multirow{5}{*}{KG-NAMM} & KEN            & ×          & 55.575         & 9.557          & 187.591         & 21.505         & -0.826         & 0.295          & 7.552          & 0.702          & 2.229          & 0.165          & -0.675         & 0.464          \\
 & MrAP           & ×          & 26.589         & 1.930          & 141.571         & 6.451          & -0.016         & 0.018          & 4.910          & 0.840          & 1.941          & 0.200          & -0.473         & 1.147          \\
 & LiteralE       & ×          & 23.783         & 0.427          & 125.736         & 4.110          & 0.167          & 0.066          & 4.471          & 1.068          & 1.744          & 0.218          & 0.064          & 0.281          \\
 & mkbe           & ×          & 28.314         & 1.913          & 145.974         & 7.222          & -0.017         & 0.053          & 4.572          & 1.125          & 1.716          & 0.316          & -0.102         & 0.099          \\
 & TransEA        & ×          & 33.386         & 2.922          & 158.248         & 6.959          & -0.186         & 0.098          & 5.270          & 1.032          & 1.858          & 0.262          & -0.200         & 0.173          \\
\hline
\multirow{1}{*}{KG-AM} & KANO       & \checkmark          & 4.753          & 1.070          & 50.605          & 4.782          & 0.839          & 0.032          & 2.746          & 1.371          & 1.216          & 0.193          & 0.420          & 0.223          \\
\hline
\multirow{1}{*}{-} & \textbf{NR-KG(Ours)}          & \checkmark          & \textbf{3.520} & \textbf{0.931} & \textbf{42.962} & \textbf{4.389} & \textbf{0.881} & \textbf{0.029} & \textbf{2.210} & \textbf{0.949} & \textbf{1.122} & \textbf{0.164} & \textbf{0.533} & \textbf{0.152} \\
\hline

\hline
\end{tabular}
}
\end{threeparttable}
\end{table*}

\begin{figure*}[t]
\centering
\includegraphics[width=2\columnwidth]{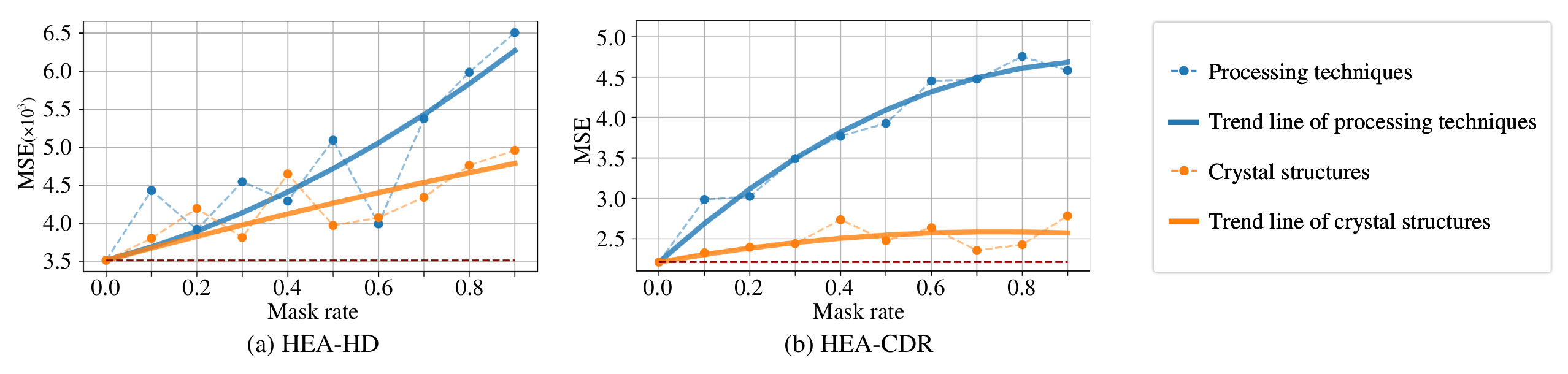} 
\caption{Semantic edges masking experimental results. (a) and (b) represent experimental results on HEA-HD and HEA-CRD datasets, respectively. As the masking ratio of semantic edges increases, the predictive MSE of NR-KG shows an upward trend, highlighting the significance of semantic information.}
\label{fig4}
\end{figure*}

\subsubsection{Baselines}
We evaluate the performance of NR-KG and compare it with the following four categories of methods:

\begin{itemize}
    \item Classical ML-based Methods (CMMs): YSRF \cite{bhandari2021yield}, HDSVR \cite{wen2019machine}, and PCHMLP \cite{bakr2022prediction, chang2019prediction}. We also verified the effectiveness of these methods when using one-hot encoding to model semantic information.
    \item  Incorporating Element Knowledge Method (IEKM): DS-HEA \cite{zhang2022composition}. 
    \item KG Numerical Attribute Modeling Methods (KG-NAMMs): KEN \cite{cvetkov2023relational}, MrAP \cite{bayram2021node}, LiteralE \cite{kristiadi2019incorporating}, mkbe \cite{pezeshkpour2018embedding} and TransEA \cite{wu2018knowledge}.
    \item Two-stage Pretrained KG Augmentation Method (KG-AM): KANO \cite{fang2023knowledge}.
\end{itemize}

\subsubsection{Experimental Results}
As shown in Table \ref{mianTable}, the results demonstrate that NR-KG achieved the best performance across all evaluation metrics. Specifically, NR-KG achieved a 25.9\% and 16.1\% improvement over the second-best method in terms of MSE on the HEA-HD and HEA-CRD, respectively.

The experimental results show that under the constraint of limited small-sample data, CMMs exhibit proficiency in capturing numerical features. Upon incorporating semantic information through one-hot encoding, regression accuracy improves, indicating a correlation between semantic information and material properties. 
IEKM introduces additional elemental knowledge but also significant redundancy in predicting HEA properties, considering only seven types of elements, and lacks sufficient constraint information from small-sample data. 
The numerical modeling of KG-NAMMs is semantic-oriented and exhibits poor modeling performance for material data with highly nonlinear features. Moreover, these methods mostly cannot introduce new nodes during the testing phase. To enable normal training and testing, although new nodes' semantic information was included during the training of the KG-NAMMs method to reduce task difficulty, the performance in predicting material properties is still very unsatisfactory. This indicates that merely aligning numerical values to the KG space cannot effectively handle the gap between numerical information and semantic information.
While KG-AM shows potential, it integrates semantic information into the numerical prediction process through a pretrained KG. However, the two-stage method lacks interaction between numerical and semantic information, limiting its performance.
As an end-to-end approach, \textbf{NR-KG} simultaneously mines numerical and semantic information, effectively leveraging data content, and achieves state-of-the-art performance.

\subsection{Ablation Study\label{E.4}}

\subsubsection{Integrity of Semantic Information} 
This experiment demonstrates NR-KG's predictive accuracy under varying percentages of missing semantic nodes. During the training stage, we randomly masked a certain percentage of HEA nodes' semantic edges related to processing techniques or crystal structures to simulate incomplete semantic information.
Fig. \ref{fig4} shows the trend of declining predictive accuracy for NR-KG on both HEA datasets as semantic information is absent. Fluctuations might stem from randomness introduced by the small dataset size. This observation underscores the significance of incorporating semantic information, such as processing techniques and crystal structures, for accurate material property prediction. The greater the information's comprehensiveness, the more effectively NR-KG leverages it to enhance predictive performance.

\begin{table*}[t]
\caption{Experimental results on the effectiveness of model structure design. Results are shown as “mean ± standard deviation”. Bold cells highlight the best results per metric.}
\label{abTable1}
\centering
\begin{threeparttable}
\resizebox{\linewidth}{!}{
\centering
\renewcommand\tabcolsep{6pt}
\begin{tabular}{@{}c|c|c | r@{±}l r@{±}l r@{±}l | r@{±}l r@{±}l r@{±}l@{}}
\hline

\hline
\multicolumn{3}{c|}{Datasets}              & \multicolumn{6}{c|}{HEA-HD}                                                             & \multicolumn{6}{c}{HEA-CRD}                                              \\
\hline
\multicolumn{3}{c|}{Task}                 & \multicolumn{6}{c|}{Regression (Testing set)}                                                            & \multicolumn{6}{c}{Regression   (Testing set)}                                                         \\
\hline
Methods & Module & Training & \multicolumn{2}{c}{MSE(×10$^3$)$\downarrow$} & \multicolumn{2}{c}{MAE$\downarrow$} & \multicolumn{2}{c|}{R$^2$$\uparrow$} & \multicolumn{2}{c}{MSE$\downarrow$} & \multicolumn{2}{c}{MAE$\downarrow$} & \multicolumn{2}{c}{R$^2$$\uparrow$}  \\
\hline
MLP       &   MLP base      &  \multirow{2}{*}{end-to-end}  & 6.017          & 1.976          & 58.110          & 9.069          & 0.801          & 0.046          & 6.633          & 1.437          & 2.125          & 0.220          & -0.426         & 0.304          \\
MLP       & +semantic        &           & 5.557          & 1.483          & 55.417          & 7.371          & 0.810          & 0.057          & 2.633          & 0.986          & 1.161          & 0.146          & 0.449          & 0.168          \\
\hline
normal-KG   & KG base          & \multirow{3}{*}{two-stage}    & 33.264         & 3.963          & 148.058         & 9.167          & -0.129         & 0.166          & 6.519          & 1.028          & 2.086          & 0.180          & -0.435         & 0.400          \\
NR-KG-MLP & +numerical value &      & 15.859         & 7.708          & 103.799         & 28.771         & 0.436          & 0.343          & 4.612          & 1.300          & 1.772          & 0.284          & 0.039          & 0.111          \\
NR-KG-GCN & +GCN             &      & 10.636         & 2.609          & 77.610          & 13.947         & 0.638          & 0.100          & 2.731          & 1.207          & 1.224          & 0.164          & 0.431          & 0.180          \\
\hline
normal-KG   & KG base          & \multirow{3}{*}{end-to-end} & 32.926         & 6.386          & 149.758         & 16.358         & -0.103         & 0.091          & 6.145          & 1.299          & 2.056          & 0.219          & -0.312         & 0.226          \\
NR-KG-MLP & +numerical value &   & 4.535          & 1.236          & 49.187          & 7.529          & 0.849          & 0.032          & 3.657          & 1.269          & 1.529          & 0.304          & 0.247          & 0.176          \\
NR-KG-GCN & +GCN             &   & \textbf{3.520} & \textbf{0.931} & \textbf{42.962} & \textbf{4.389} & \textbf{0.881} & \textbf{0.029} & \textbf{2.210} & \textbf{0.949} & \textbf{1.122} & \textbf{0.164} & \textbf{0.533} & \textbf{0.152}  \\
\hline

\hline
\end{tabular}
}

\end{threeparttable}
\end{table*}

\begin{table*}[t]
\caption{Experimental results on effect of $\mathcal{L}_C$ and $\mathcal{L_D}$. Results are presented as the mean MSE from 6-fold cross-validation. Bold cells indicate the best results in each row.}
\label{abTable2}
\centering
\begin{threeparttable}
\resizebox{\linewidth}{!}{
\centering
\renewcommand\tabcolsep{8pt}
\begin{tabular}{@{} c|c|c|ccccccccccc @{}}
\hline

\hline
Datasets & \multicolumn{2}{c|}{$\gamma_a$}  & 0.00  & 0.04  & 0.08  & 0.12  & 0.16           & 0.20  & 0.30  & 0.40  & 0.60  & 0.80  & 1.00  \\
\hline
HEA-HD     & MSE(×10$^3$)$\downarrow$  & \multirow{2}{*}{$\gamma_b$=0.00} & 4.108 & 4.038 & 3.853 & 3.794 & \textbf{3.620} & 3.780 & 3.787 & 3.840 & 3.908 & 3.902 & 3.944 \\
HEA-CRD    & MSE$\downarrow$    &                            & 2.681 & 2.400 & 2.343 & 2.324 & \textbf{2.303} & 2.335 & 2.346 & 2.377 & 2.455 & 2.490 & 2.555 \\
\hline

\hline
\multicolumn{1}{c|}{Datasets} & \multicolumn{2}{c|}{$\gamma_b$}               & 0.00  & 0.02           & 0.04           & 0.06  & 0.08  & 0.10  & 0.12  & 0.14  & 0.16  & 0.18  & 0.20  \\
\hline
HEA-HD          & MSE(×10$^3$)$\downarrow$ & \multirow{2}{*}{$\gamma_a$=0.16} & 3.620 & 3.567          & \textbf{3.520} & 3.728 & 4.081 & 4.189 & 4.173 & 4.170 & 4.204 & 4.430 & 4.471 \\
HEA-CRD         & MSE$\downarrow$     &                            & 2.303 & \textbf{2.210} & 2.498          & 2.603 & 2.634 & 2.699 & 2.678 & 2.691 & 2.745 & 2.747 & 2.873 \\
\hline

\hline
\end{tabular}
}

\end{threeparttable}
\end{table*}

\subsubsection{Ablation Analysis of Components}
We explored the impact of different components of the NR-KG approach. A baseline reference was established using MLP, a neural network-based numerical modeling method. 
For the ``normal-KG'' approach, a normal KG was constructed solely using semantic information, and TransE \cite{bordes2013translating} was employed for representation learning.
In the NR-KG-MLP variant, the GCN in NR-KG was replaced with MLP. 
The complete NR-KG method encompassing all components was also evaluated, referred to as NR-KG-GCN for differentiation.
Both two-stage and end-to-end learning were applied to the ``normal-KG'' and NR-KG-based methods, respectively. 

The results in Table \ref{abTable1} clearly demonstrate the limitations of the ``normal-KG'' approach, which relies solely on semantics for property prediction and emphasizes the crucial role of numerical data in achieving accurate predictions.
The inclusion of numerical data significantly enhances NR-KG-MLP compared to the ``normal-KG'' approach, underscoring the effectiveness of the cross-modal KG and canonical KG.
Furthermore, the incorporation of GCN further improves predictive performance, highlighting the significance of interactions between HEA nodes and their semantic neighbors.
Significantly, the end-to-end approach consistently outperforms the two-stage method by leveraging the synergy between KG representation learning and numerical prediction. This approach efficiently learns projection spaces that integrate numerical and semantic aspects, capturing relations within the samples.

\subsubsection{Comparative Learning Validation} 
As the design of the projection prediction loss (PPL) considers the inverse process of the comparative learning loss (CLL), we initially adjust the hyperparameter $\gamma_a$ in $\mathcal{L}$ and subsequently adjust $\gamma_b$ to evaluate their impact on model performance.
The results in Table \ref{abTable2} show a performance trend where increasing $\gamma_a$ initially enhances the model's performance, followed by a decline. 
Optimal performance for two HEA datasets occurs at $\gamma_a\!=\!0.16$. 
Similarly, with $\gamma_a\!=\!0.16$, changing $\gamma_b$ yields similar performance trends, with HEA-HD and HEA-CRD reaching their peaks at $\gamma_b\!=\!0.04$ and $\gamma_b\!=\!0.02$ respectively.
These findings highlight that proper PPL and CCL supervision in canonical KG learning can enhance regression tasks.

\subsection{Interpretability of NR-KG\label{E.5}} 

\subsubsection{Visualization of Semantic Nodes in Canonical KG}
\begin{figure*}[t]  
    \includegraphics[width=2\columnwidth]{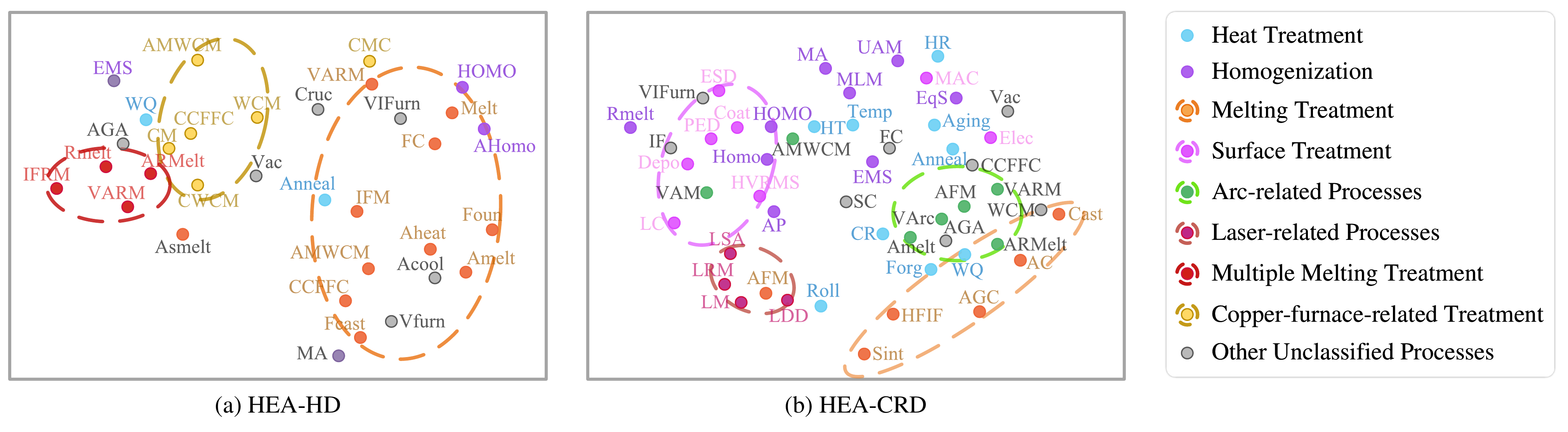} 
    \caption{t-SNE visualization of processing technique vectors in the canonical KG, where each point represents a technique. The abbreviation list is in Table \ref{TabAI}.}
    \label{fig5}
\end{figure*}

We utilized t-SNE dimensionality reduction to visualize the vectors of processing technique nodes in the canonical KG. Nodes were colored based on the classification of processing technique vocabulary by domain experts in the reference materials field, as shown in Fig. \ref{fig5}. Some nodes exhibit spatial patterns consistent with material science expectations. Despite the limited disambiguation and alignment during dataset construction, NR-KG successfully extracts semantic insights from known semantic and numerical information. 
This potentially explains the enhanced predictive capability of NR-KG when incorporating processing technique data.

\begin{table*}[!t]
\caption{Interpretability of crystal structure nodes vectors and proxy nodes vectors in canonical KG. Results are shown as ``mean ± standard deviation''. Bold cells highlight the best results per metric. ``NR-KG($\mathcal{L}_M=0$)'' configures NR-KG training to rely solely on $\mathcal{L}_C$ and $\mathcal{L}_D$, transforming it into a link prediction method.}
\label{abTable3}
\centering
\resizebox{\linewidth}{!}{
\begin{threeparttable}
\centering
\renewcommand\tabcolsep{14pt}
\begin{tabular}{c|ccc|ccc}
\hline

\hline
Datasets                      & \multicolumn{3}{c|}{HEA-HD}                           & \multicolumn{3}{c}{HEA-CRD}               \\
\hline
Task                         & \multicolumn{3}{c|}{Link prediction (Testing set)}                     & \multicolumn{3}{c}{Link prediction (Testing set)}                     \\
\hline
Methods                      & MRR$\uparrow$         & MR$\downarrow$                & Hit@1$\uparrow$                & MRR$\uparrow$                  & MR$\downarrow$                & Hit@1$\uparrow$                \\
\hline
Uniform sampling random      & 0.472±0.031          &   3.125±0.233    & 0.222±0.039          & 0.395±0.011          &  3.872±0.105  & 0.154±0.017          \\
Distribution sampling random & 0.792±0.016          &  1.478±0.036  & 0.614±0.031          & 0.804±0.013          &  1.550±0.044 & 0.656±0.025          \\
NR-KG($\mathcal{L}_M=0$)             & 0.932±0.019 & 1.143±0.041 & 0.868±0.038 & 0.885±0.081 & 1.375±0.264 & 0.809±0.137 \\
NR-KG     & \textbf{0.934±0.011}   & \textbf{1.142±0.027} & \textbf{0.874±0.020} & \textbf{0.939±0.006}    & \textbf{1.251±0.050} & \textbf{0.911±0.007}  \\
\hline

\hline

\end{tabular}
\end{threeparttable}
}
\end{table*}

\subsubsection{Crystal Structure Link Completion}
We employed the vectors from the canonical KG to complete the links between HEA nodes and crystal structure nodes, and compared the results with two random methods. The results, presented in Table \ref{abTable3}, reveal that even though our model primarily focuses on material property prediction, it can effectively capture the semantic information of nodes. The accuracy in predicting crystal structures reaches 87.4\% and 91.1\% in HEA-HD and HEA-CRD. 
Moreover, in comparison with NR-KG ($\mathcal{L}_M=0$), we observed that learning material properties enhances the model's ability to comprehend semantics. 
This observation underscores NR-KG's role in bridging the semantic-numerical gap and effectively modeling relationships between samples, thereby extracting more information from small-sample data. 
The results verify NR-KG's capability to bridge the semantic-numerical gap and capture inter-sample relationships.
This experiment aligns with the empirical understanding in the material field that ``process influences structure and structure influences properties'' \cite{luo2020strong}, ultimately contributing to the remarkable performance of NR-KG.

\begin{table}[!t]
\caption{Molecular experiment results. Results are shown as ``mean ± standard deviation''. Bold cells highlight the best results per metric. Our method, NR-KG, achieved a 54.3\% and 22.2\% improvement over the second-best algorithm of RMSE. This demonstrates the strong performance of NR-KG on public datasets.}
\label{asTable1}
\centering
\renewcommand\tabcolsep{20pt}
\resizebox{\linewidth}{!}{
\begin{threeparttable}
\begin{tabular}{c|r@{±}l|r@{±}l}
\hline

\hline
Datasets       & \multicolumn{2}{c|}{FreeSolv}  & \multicolumn{2}{c}{ESOL} \\
\hline
No. molecules & \multicolumn{2}{c|}{642} & \multicolumn{2}{c}{1,128}      \\
\hline
Task          & \multicolumn{4}{c}{Regression   (Testing set)}           \\
\hline
Methods       & \multicolumn{4}{c}{RMSE$\downarrow$}                                 \\
\hline
GCN & 2.870 & 0.135 & 1.431 & 0.050 \\
GIN & 2.765 & 0.180 & 1.452 & 0.020 \\
MPNN & 1.621 & 0.952 & 1.167 & 0.430 \\
DMPNN & 1.673 & 0.082 & 1.050 & 0.008 \\
CMPNN & 1.570 & 0.442 & 0.798 & 0.112 \\
AttentiveFP & 2.073 & 0.183 & 0.877 & 0.029 \\
N-Gram$_{\rm RF}$ & 2.688 & 0.085 & 1.074 & 0.107 \\
N-Gram$_{\rm XGB}$ & 5.061 & 0.744 & 1.083 & 0.082 \\
PretrainGNN & 2.764 & 0.002 & 1.100 & 0.006 \\
GROVER$_{\rm base}$ & 2.176 & 0.052 & 0.983 & 0.090 \\
GROVER$_{\rm large}$ & 2.272 & 0.051 & 0.895 & 0.017 \\
MolCLR & 2.594 & 0.249 & 1.271 & 0.040 \\
GEM & 1.877 & 0.094 & 0.798 & 0.029 \\
Uni-Mol & 1.620 & 0.035 & 0.788 & 0.029 \\
KANO & 1.142 & 0.258 & 0.670 & 0.019 \\
\hline
\textbf{NR-KG(Ours)} & \textbf{0.584} & \textbf{0.235} & \textbf{0.521} & \textbf{0.042} \\
\hline

\hline

\end{tabular}

\end{threeparttable}
}
\end{table}

\subsection{Application Scalability Experiment\label{E.6}}
To validate the application potential of NR-KG on scientific data, we used publicly available molecular datasets with graph structures on the field of physical chemistry to verify NR-KG. 

\subsubsection{Baselines}
To demonstrate the generalization ability of NR-KG, we comprehensively evaluate its performance on an open dataset in the field of physical chemistry in comparison with molecular-graph-based and molecular-graph pre-training baselines, which encompass the currently recognized state-of-the-art methods, which include GCN \cite{kipf2016semi}, GIN \cite{xu2018powerful}, MPNN \cite{gilmer2017neural}, DMPNN \cite{yang2019learned}, CMPNN \cite{song2020communicative}, AttentiveFP \cite{xiong2019pushing}, N-Gram$_{\rm RF}$ \cite{liu2019n}, N-Gram$_{\rm XGB}$ \cite{liu2019n}, PretrainGNN \cite{hu2019strategies}, GROVER$_{\rm base}$ \cite{rong2020self}, GROVER$_{\rm large}$ \cite{rong2020self}, MolCLR \cite{wang2022molecular}, GEM \cite{fang2022geometry}, Uni-Mol \cite{zhou2023uni}, KANO \cite{fang2023knowledge}.
For all the baseline methods we report the results taken from the referred papers \cite{fang2022geometry, zhou2023uni, fang2023knowledge}, as shown in Table \ref{asTable1}.

\subsubsection{Experiment Results}
The results in Table \ref{asTable1} show that NR-KG outperforms other methods, achieving state-of-the-art performance on molecular datasets. Specifically, on the ESOL dataset and FreeSolv dataset, NR-KG exhibits improvements in RMSE by 22.2\% and 54.3\%, respectively, compared to the second-best method.
Despite not being specifically designed for molecular tasks, NR-KG achieves superior performance by incorporating a simple adaptation for molecular graph-type data through the addition of a graph feature encoder. 
It considers molecular fingerprints as semantic information to enhance substructures contained in the molecules, surpassing existing methods in predictive accuracy. This highlights NR-KG's ability to effectively utilize data from small samples, capturing relevant information from molecular graphs and fingerprints to make accurate predictions.

In summary, the experimental results indicate that the NR-KG model is not only suitable for alloy material datasets but also for predicting molecular properties, showcasing its potential for various types of data.

\section{Limitations and Discussion}

\subsection{Limitations}
Although NR-KG effectively utilizes the numerical-semantic information of small-sample data and inter-sample relationship information, yielding promising results in material property prediction tasks, the method's application is constrained in large-scale datasets due to limitations in the graph convolutional network (GCN) stage. In each iteration, the feature processing of the complete cross-modal KG needs to be considered, leading to increased memory requirements as the dataset and cross-modal KG size grow. This hinders the applicability of NR-KG in large-scale datasets.

\subsection{Impact and Prospects}

NR-KG provides an innovative solution for integrating numerical and semantic data, introducing new perspectives and tools to the community of machine learning (ML) and scientific data.

In the realm of ML, particularly in KGs and small-sample learning domains:
\begin{itemize}
    \item NR-KG mines data features and captures information and inter-sample correlations within limited data samples, providing a novel approach for researchers in the field of small-sample ML.
    \item NR-KG opens up potential avenues for handling more complex modalities and heterogeneous scientific data, expanding the application scope of knowledge graph technology in vertical domains.
    \item NR-KG has demonstrated potential applications across various scientific domains. In the future, we will further explore the characteristics of data in different scientific domains, aiming to enhance NR-KG to become a more flexible and efficient ML algorithm, contributing to AI for science and interdisciplinary research.
\end{itemize}

In the community of scientific data:
\begin{itemize}
    \item In the field of materials science, we propose and publicly release two new High-Entropy Alloy (HEA) datasets, aiming to foster the scientific open-source community and promote the development of high-quality scientific datasets. This work also encourages the collection of diverse datasets in various scientific fields.
    \item The NR-KG paradigm provides new tools for research in the materials science field. Its excellent interpretability aids in uncovering the complex relationships behind material properties, facilitating in-depth research into phenomena not yet fully understood in materials.
\end{itemize}

As the first end-to-end KG numerical reasoning method, NR-KG is expected to inspire other tasks in ML, such as cross-modal or small-sample learning. 
We anticipate that the principles of this algorithm can enhance task performance and provide interpretability to the results. For instance, in common-sense knowledge graph link prediction tasks, more rational embeddings can be obtained through the interaction between numerical and semantic information.
In the scientific domain, we hope that NR-KG becomes a valuable tool for scientists, enhancing the accuracy of ML predictions under limited samples and expediting new developments in areas such as the discovery of novel materials.

\section{Conclusion}
Material property prediction plays a crucial role in various key materials science research areas, including studies on material mechanisms and the discovery of new materials.
In this research, we proposed a novel numerical reasoning method of cross-modal KG, NR-KG, for predicting material properties. NR-KG constructs a cross-modal KG to represent material data, effectively capturing both numerical and semantic information. The NR-KG enables end-to-end numerical reasoning within the cross-modal KG.
Our proposed NR-KG method outperforms other approaches in property prediction experiments, offering a novel avenue for modeling small-sample scientific data. NR-KG demonstrates its ability to learn material informatics principles effectively even with limited samples, expanding its potential for innovative applications in the scientific domain.

\section*{CRediT authorship contribution statement}

\textbf{Guangxuan Song:} Writing - Original Draft, Software, Data Curation, Methodology. \textbf{Dongmei Fu}: Conceptualization, Supervision. \textbf{Zhongwei Qiu}: Writing - Review \& Editing. \textbf{Zijiang Yang}: Visualization. \textbf{Jiaxin Dai}: Validation. \textbf{Lingwei Ma}: Data Curation. \textbf{Dawei Zhang}: Data Curation. 

\section*{Declaration of competing interest}
The authors declare that they have no known competing financial interests or personal relationships that could have appeared to influence the work reported in this paper.

\section*{Code and Datasets Available\label{A1}}
The code and HEA datasets will be open-sourced on GitHub.

\section*{Acknowledgments}
This work was supported by the National Environmental Corrosion Platform of China. We extend our special thanks to the doctoral researchers from Prof. Dawei Zhang's team at the National Materials Corrosion and Protection Data Center, University of Science and Technology Beijing, for their assistance in data collection.

\renewcommand{\thetable}{A\arabic{table}}
\renewcommand{\thefigure}{A\arabic{figure}}
\setcounter{figure}{0}
\setcounter{table}{0}

\appendix

\section{Supplementary Material}

Table \ref{TabAI} serves as the abbreviation list for Fig. \ref{fig5} in the main text.
\begin{table*}[b]
\caption{Abbreviation of processing techniques. ``Abbr.'' means ``Abbreviation''.}
\label{TabAI}
\centering
\resizebox{\linewidth}{!}{
\begin{tabular}{ll | ll | ll}
\hline

\hline
Abbr. & Processing techniques                        & Abbr. & Processing techniques                   & Abbr. & Processing techniques                \\
\hline

\hline
Aging  & Aging                                   & Cruc  & Crucible                                 & LRM    & Laser re-melting              \\
AC     & Air casting                             & Depo  & Deposition                               & LSA    & Laser surface alloying        \\
Acool  & Air cooling                             & EMS   & Electromagnetic stirring                 & MLM    & Magnetic levitation melting   \\
AHomo  & Alloy homogenization                    & Elec  & Electroplating                           & MA     & Mechanical alloying           \\
Anneal & Annealing                               & ESD   & Electrospark deposition                  & Melt   & Melting                       \\
Arc    & Arc                                     & EqS   & Equilibrium state                        & MAC    & Microplasma arc cladding      \\
AFM    & Arc furnace melting                     & Forg  & Forging                                  & PED    & Pulsed electric deposition    \\
Afus   & Arc fusion                              & Foun  & Foundry                                  & Rmelt  & Re-melting                    \\
Aheat  & Arc heating                             & FC    & Furnace cooling                          & Roll   & Rolling                       \\
Amelt  & Arc melting                             & Fcast & Fusion casting                           & Sint   & Sintering                     \\
AMWCM  & Arc melting in water-cooled copper mold & HT    & Heat treatment                           & SC     & Slow cooling                  \\
ARMelt & Arc re-melting                          & HFIF  & High-frequency induction heating furnace & Temp   & Tempering                     \\
Asmelt & Arc smelting                            & HOMO  & Homogeneity                              & UAM    & Ultrasonic atomization method \\
AGA    & Argon gas atmosphere                    & Homo  & Homogenization                           & Vac    & Vacuum                        \\
AGC    & Argon gas casting                       & HR    & Hot rolling                              & VArc   & Vacuum arc                    \\
AP           & Atomization process                          & HVRMS        & HV radio frequency magnetron sputtering & VAM          & Vacuum arc melting                   \\
Cast   & Casting                                 & IF    & Induction furnace                        & VARM   & Vacuum arc re-melting         \\
CWCM   & Casting with copper molds               & IFM   & Induction furnace melting                & Vfurn  & Vacuum furnace                \\
Coat   & Coating                                 & IFRM  & Induction furnace re-melting             & VIFurn & Vacuum induction furnace      \\
CR     & Cold rolling                            & LC    & Laser cladding                           & WQ     & Water quenching               \\
CCFFC        & Copper crucible furnace with furnace cooling & LDD          & Laser direct deposition                 & WCCF         & Water-cooled copper crucible furnace \\
CM     & Copper mold                             & LM    & Laser melting                            & WCM    & Water-cooled copper mold      \\
CMC    & Copper mold casting                     &       &                                          &        &                              \\
\hline

\hline
\end{tabular}
}
\end{table*}

\subsection{Experiment Details\label{AE}}

\subsubsection{Implemental Details}
In \ref{E.4}, we examine the influence of two critical hyperparameters, $\gamma_a$ and $\gamma_b$, on NR-KG. The hyperparameter $\gamma$ in Eq.(\ref{eq:lc}) serves to ensure that negative samples do not constantly yield ineffective loss for $\mathcal{L}_C$. Given the normalization of $\mathbf{e}_i$ in Eq.(\ref{eq:semd}) and $\mathbf{e}_m$ in Eq.(\ref{eq:npl}), $\gamma=1$ is suggested as a versatile default.

We set the dimension of the canonical KG vectors in NR-KG as $H=128$, and the graph convolutional network (GCN) has 2 layers to prevent excessive smoothing. We applied $\rm{dropout}=0.5$ in the Property Decoder to mitigate overfitting and utilized the LeakyReLU activation function. 
The training process consisted of 2000 epochs using the Adam optimizer, and we employed the StepLR learning rate adjustment strategy with $\rm{gamma}=0.5$ and $\rm{step\_size}=1$. An early stopping strategy was implemented to enhance generalization, training stopped if the validation loss $\mathcal{L}$ did not decrease for 50 consecutive steps.

For training on the HEA-HD dataset, a learning rate of 0.005 was used with $\gamma=1$, $\gamma_a=0.16$, and $\gamma_b=0.04$ for $\mathcal{L}$. On the HEA-CRD dataset, a learning rate of 0.001 was employed with $\gamma=1$, $\gamma_a=0.16$, and $\gamma_b=0.02$. The experiments were conducted on an NVIDIA GTX 1080Ti.

For training on the molecule property datasets, we utilize the DeepChem \cite{Ramsundar-et-al-2019} to initialize molecular feature graphs. We adjust learning rates from 0.001, $\gamma_a=0.2$ and $\gamma_b=0.04$ for $\mathcal{L}$. All other hyperparameters remained consistent with the experiments outlined on the HEA dataset. The experiments were carried out on an NVIDIA GTX 4090.

\begin{table}[t!]
\caption{Resource utilization in experiments. Results are reported as ``mean ± standard deviation''. For the HEA-HD and HEA-CRD datasets, 6-fold cross-validation was used, whereas for FreeSolv and ESOL, results are based on three independent data splits, consistent with Sec. \ref{E.3} and \ref{E.6}, respectively.}
\label{Tab:resource}
\centering
\resizebox{\linewidth}{!}{
\begin{tabular}{ c|cc|c }
\hline

\hline
Datasets & \multicolumn{2}{c|}{Metric} & NR-KG \\ 
\hline

\hline
\multirow{5}{*}{HEA-HD} & \multicolumn{2}{c|}{Parameters} & 0.074 M \\ \cline{2-4} 
 & \multicolumn{1}{c|}{\multirow{2}{*}{Training}} & Total time (Training set size = 214) & 13.92 ± 3.33 s \\ \cline{3-4} 
 & \multicolumn{1}{c|}{} & Max GPU memory usage & 1.11 ± 0.00 G \\ \cline{2-4} 
 & \multicolumn{1}{c|}{\multirow{2}{*}{Inference}} & Total time (Testing set size = 53) & 0.04 ± 0.00 s \\ \cline{3-4} 
 & \multicolumn{1}{c|}{} & Max GPU memory usage & 1.12 ± 0.01 G \\ 
\hline

\hline
\multirow{5}{*}{HEA-CRD} & \multicolumn{2}{c|}{Parameters} & 0.077 M \\ \cline{2-4} 
 & \multicolumn{1}{c|}{\multirow{2}{*}{Training}} & Total time (Training set size = 92) & 8.09 ± 1.59 s \\ \cline{3-4} 
 & \multicolumn{1}{c|}{} & Max GPU memory usage & 0.27 ± 0.00 G \\ \cline{2-4} 
 & \multicolumn{1}{c|}{\multirow{2}{*}{Inference}} & Total time (Testing set size = 25) & 0.01 ± 0.00 s \\ \cline{3-4} 
 & \multicolumn{1}{c|}{} & Max GPU memory usage & 0.24 ± 0.00 G \\ 
\hline

\hline
\multirow{5}{*}{FreeSolv} & \multicolumn{2}{c|}{Parameters} & 0.088 M \\ \cline{2-4} 
 & \multicolumn{1}{c|}{\multirow{2}{*}{Training}} & Total time (Training set size = 513) & 72.59 ± 10.79 s \\ \cline{3-4} 
 & \multicolumn{1}{c|}{} & Max GPU memory usage & 2.99 ± 0.24 G \\ \cline{2-4} 
 & \multicolumn{1}{c|}{\multirow{2}{*}{Inference}} & Total time (Testing set size = 65) & 0.11 ± 0.00 s \\ \cline{3-4} 
 & \multicolumn{1}{c|}{} & Max GPU memory usage & 2.99 ± 0.24 G \\ 
\hline

\hline
\multirow{5}{*}{ESOL} & \multicolumn{2}{c|}{Parameters} & 0.088 M \\ \cline{2-4} 
 & \multicolumn{1}{c|}{\multirow{2}{*}{Training}} & Total time (Training set size = 902) & 133.86 ± 9.82 s \\ \cline{3-4} 
 & \multicolumn{1}{c|}{} & Max GPU memory usage & 4.76 ± 0.24 G \\ \cline{2-4} 
 & \multicolumn{1}{c|}{\multirow{2}{*}{Inference}} & Total time (Testing set size = 114) & 0.23 ± 0.02 s \\ \cline{3-4} 
 & \multicolumn{1}{c|}{} & Max GPU memory usage & 4.76 ± 0.24 G \\ 
\hline

\hline
\end{tabular}
}
\end{table}

\subsubsection{Implemental Details}

See Table \ref{Tab:resource} for a report on the resource consumption of the main experiments in Sec. \ref{E.3} and \ref{E.6}. By effectively utilizing numerical and semantic information, NR-KG achieves state-of-the-art (SOTA) results with a lightweight model of under 0.09M parameters. In terms of time and GPU memory usage, although material property prediction tasks typically have low requirements for real-time performance, NR-KG’s superior data utilization and construction of semantic correlations among data samples within limited data volumes obviate the need for large-scale networks. Consequently, model training can be completed in seconds or minutes, and inference on test set data can be done in milliseconds. With a maximum GPU memory usage of 4.76G, NR-KG can run on most consumer-grade GPUs when dealing with small sample sizes.

\begin{figure*}[t]
\centering
\includegraphics[width=1.9\columnwidth]{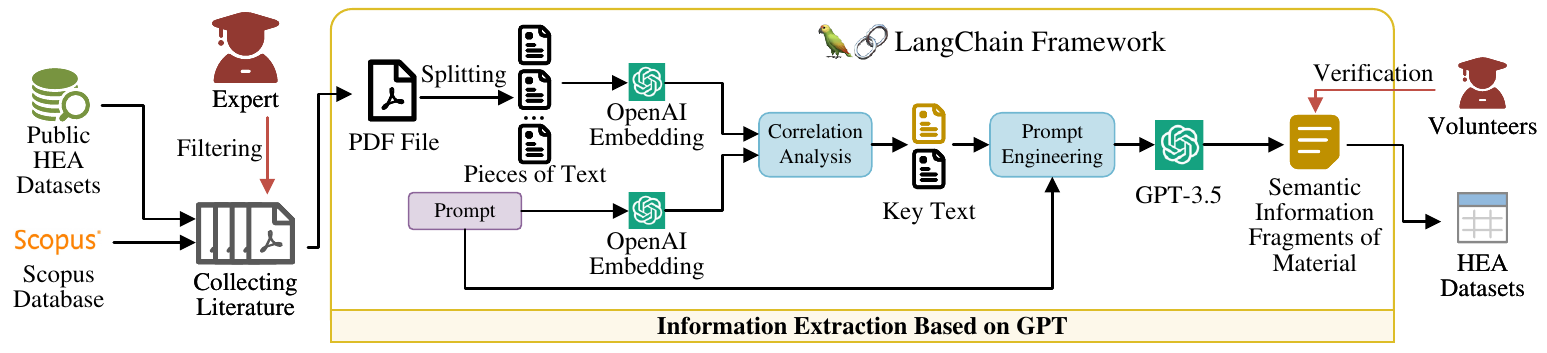} 
\vspace{-0.1cm}
    \caption{The semi-automated text extraction framework. We collected data from public HEA dataset sources and high-level papers in the field of HEA from Scopus. Using LangChain, we developed a semi-automated information extraction framework, which was then validated and documented by volunteers from the materials domain.}
    \label{fig:figA1}
\end{figure*}

\begin{figure*}[!t]
\centering
\includegraphics[width=1.9\columnwidth]{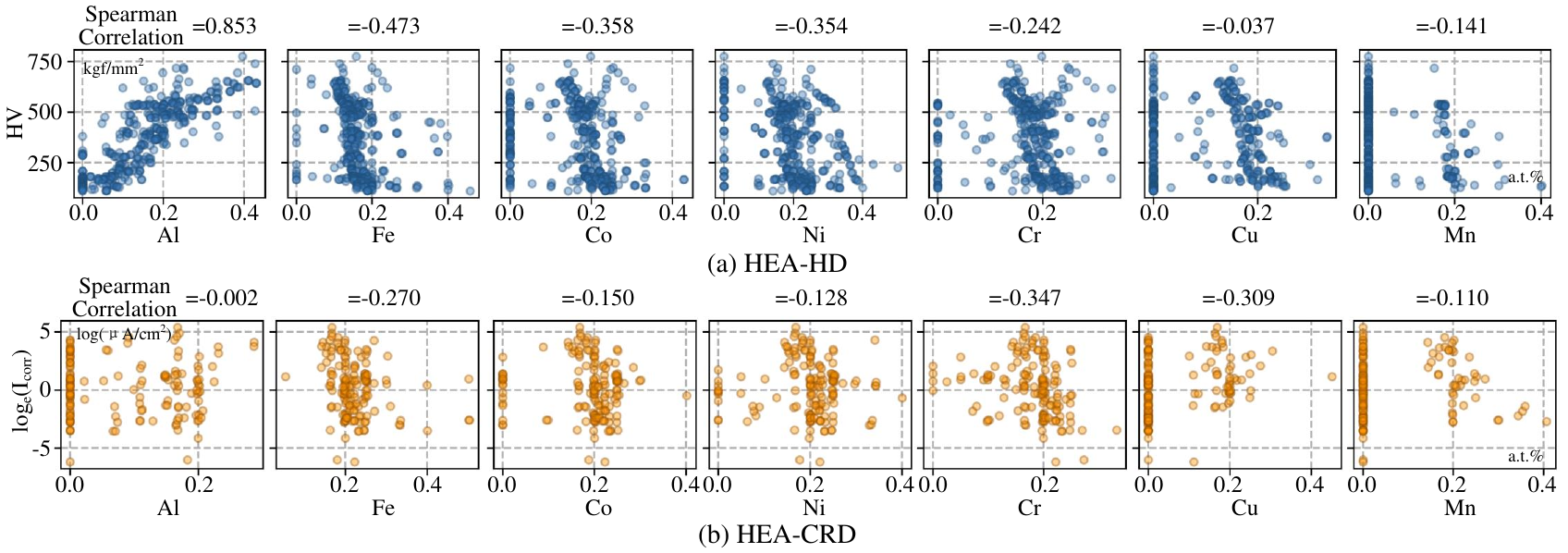} 
\vspace{-0.1cm}
    \caption{The correlation between individual element compositions and material properties.}
    \label{fig:figA2}
\vspace{-0.5cm}
\end{figure*}

\begin{figure}[!t]
\centering
\includegraphics[width=1\columnwidth]{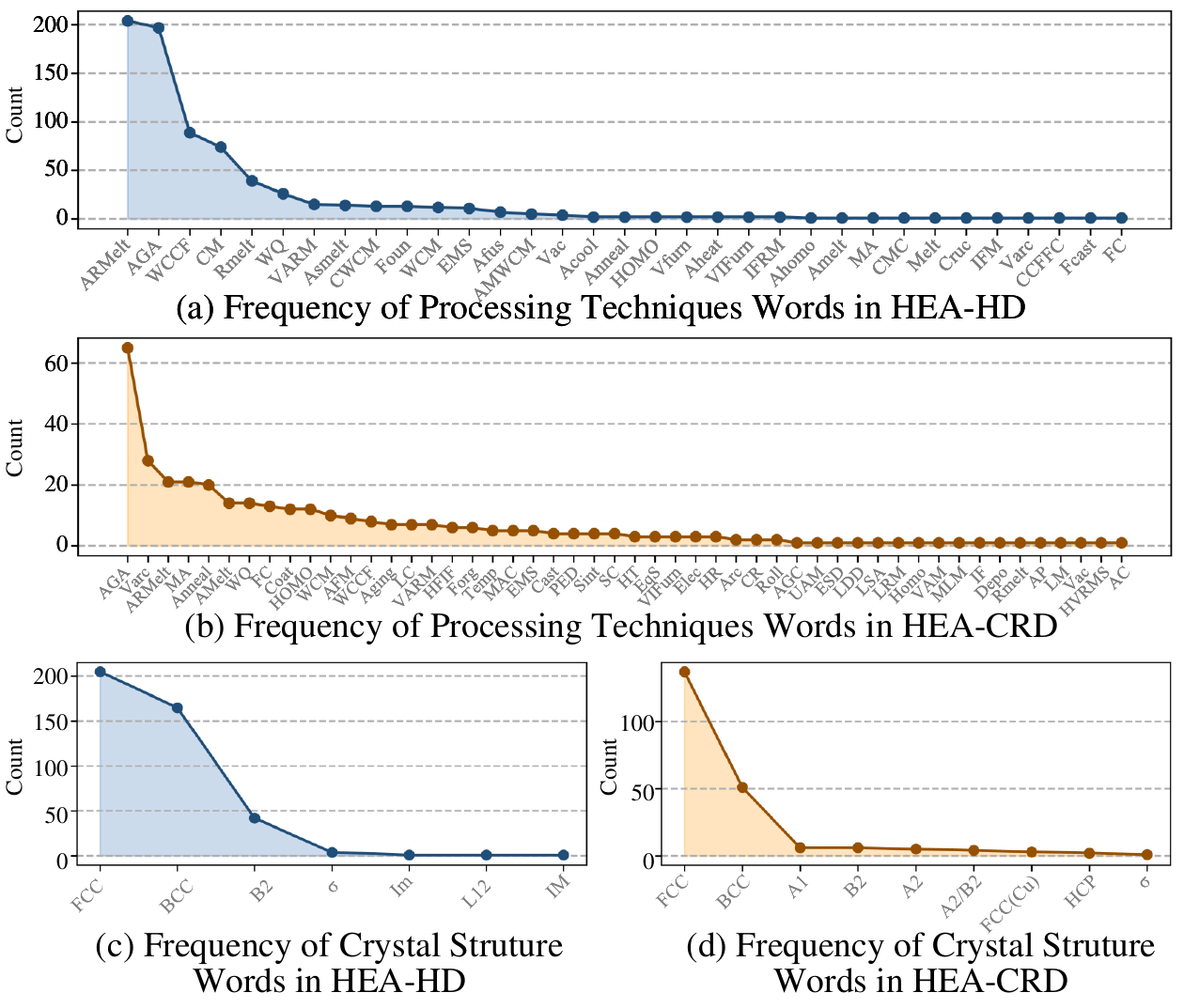} 
\caption{Frequency of words in HEA-HD and HEA-CRD.}
\label{fig:figA3}
\end{figure}

\subsection{HEA Datasets Analysis\label{A3}}

\subsubsection{Data Collection\label{apx_dataCollection}}
The data collection effort was carried out by our team and volunteers from the materials field under the guidance of domain experts. As illustrated in Fig. \ref{fig:figA1}, we initially compiled existing publicly available materials datasets from \cite{yang2022machine, nyby2021electrochemical, xiong2021machine}. We reviewed the source materials referenced during the data collection of these public datasets and extracted corresponding material processing techniques and crystal structure information. 
Furthermore, we collected HEA information reported in recent high-quality literature in the HEA field. 
For the aforementioned process, we developed a semi-automated text extraction framework by LangChain \cite{langchain}: 
we first parsed PDFs and split paragraphs, then measured the correlation between the PDF paragraphs and prompts using OpenAI Embedding \cite{gptemd} API. Subsequently, paragraphs with sufficient similarity were fed into GPT-3.5 \cite{gpt35} API after prompt engineering to extract precise sections containing descriptions of HEA processing techniques and crystal structures. Lastly, volunteers from the materials field performed verification and recording. The utilization of GPT-3.5 reduced the workload of literature reading and provided insights for building an automated literature data collection pipeline in the future.

\subsubsection{Sample Distribution Analysis}
The data collection scope encompasses HEAs with Al-Fe-Ni-Co-Cr-Cu-Mn elements. 
The HEAs we have collected consist of 3-6 elements from the set of Al-Fe-Ni-Co-Cr-Cu-Mn, totaling 7 elements. Consequently, the numerical data contains a substantial number of zero values. This prevalence of zero values is a distinctive characteristic of our HEA dataset. 

As shown in Fig. \ref{fig:figA2}, we depict the correlation between individual element compositions and material properties, along with calculated Spearman correlation coefficients. Analysis reveals that the Al element in the HEA-HD dataset exhibits a noticeable correlation with hardness, while Fe, Co, and Ni also demonstrate certain correlations. Conversely, the correlations within HEA-CRD are notably weak. These findings align with the comparative experimental results in the full papers, indicating that methods with stronger numerical modeling capabilities yield better coefficient of determination (R$^2$) results on HEA-HD than on HEA-CRD. The insufficient correlation between elements and performance in HEA-CRD underscores the presence of complex non-linear relationships between material composition and performance, posing challenges for modeling. 

\balance

Fig. \ref{fig:figA3} demonstrates a pronounced long-tail distribution in semantic information, which further contributes to modeling challenges arising from data imbalance. In summary, HEA-HD and HEA-CRD are challenging small-sample material property datasets that incorporate both numerical and semantic information.

\newpage

\bibliographystyle{unsrt}

\bibliography{citation}

\end{document}